\documentclass[journal]{IEEEtran}
\usepackage{amsmath,amsfonts,amssymb}
\usepackage{algpseudocode}
\usepackage{algorithm}
\usepackage{array}
\usepackage{comment}
\usepackage[caption=false,font=normalsize,labelfont=sf,textfont=sf]{subfig}
\usepackage{xcolor}
\usepackage[most]{tcolorbox}
\newtcolorbox[auto counter, number within=section]{takeawaybox}[2][]{colback=yellow!10!white,colframe=orange!80!black,fonttitle=\bfseries, title=Takeaway, #1}
\definecolor{finalblue}{RGB}{0,114,178}

\usepackage{booktabs}
\usepackage{tikz}

\usepackage{tcolorbox}
\tcbset{
  takeawaybox/.style={
    colback=white,
    colframe=finalblue,
    boxrule=1pt,
    arc=1pt,
    left=6pt,
    right=6pt,
    top=6pt,
    bottom=6pt,
    fonttitle=\bfseries,
  }
}

\usepackage{textcomp}
\usepackage{stfloats}
\usepackage{url}
\usepackage{verbatim}
\usepackage{graphicx}
\usepackage{cite}
\usepackage{hyperref}
\usepackage{booktabs}
\usepackage{adjustbox}
\usepackage{makecell}
\usepackage{multirow}
\usepackage{placeins}
\usepackage{tabularx}

\begin{document}

\title{Geometric Structural Knowledge Graph Foundation Model}

\author{IEEE Publication Technology,~\IEEEmembership{Staff,~IEEE,}
\thanks{This paper was produced by the IEEE Publication Technology Group. They are in Piscataway, NJ.}
\thanks{Manuscript received April 19, 2021; revised August 16, 2021.}}

\author{Ling Xin*,
        Mojtaba~Nayyeri*,
         Zahra~Makki~Nayeri,
        Steffen~Staab%
\thanks{* Equal contribution\\
Ling Xin, Mojtaba Nayyeri, and Steffen Staab are with the University of Stuttgart, Stuttgart, Germany (e-mail: mojtaba.nayyeri@ki.uni-stuttgart.de).}%
\thanks{Steffen Staab is with University of Southampton, UK (e-mail: steffen.staab@ki.uni-stuttgart.de).}%
\thanks{Zahra Makki Nayeri is with Shahrood University of Technology, Shahrood, Iran (e-mail: zmakki@shahroodut.ac.ir).}%
}

\maketitle

\begin{abstract}
Structural knowledge graph foundation models aim to generalize reasoning to completely new graphs with unseen entities and relations. 
A key limitation of existing approaches like \textsc{Ultra} is their reliance on a single relational transformation (e.g., element-wise multiplication) in message passing, which can constrain expressiveness and fail to capture diverse relational and structural patterns exhibited on diverse graphs.
In this paper, we propose \textsc{Gamma}, a novel foundation model that introduces multi-head geometric attention to knowledge graph reasoning. 
\textsc{Gamma} replaces the single relational transformation with multiple parallel ones, including real, complex, split-complex, and dual number based transformations, each designed to model different relational structures.
A relational conditioned attention fusion mechanism then adaptively fuses them at link level via a lightweight gating with entropy regularization, allowing the model to robustly emphasize the most appropriate relational bias for each triple pattern. 
We present a full formalization of these algebraic message functions and discuss how their combination increases expressiveness beyond any single space. 
Comprehensive experiments on 56 diverse knowledge graphs demonstrate that \textsc{Gamma} consistently outperforms \textsc{Ultra} in zero-shot inductive link prediction, with a 5.5\% improvement in mean reciprocal rank on the inductive benchmarks and a 4.4\% improvement across all benchmarks, highlighting benefits from complementary geometric representations.
\end{abstract}


\section{Introduction}
\IEEEPARstart{K}{nowledge} Graphs (KGs) store factual information as triples \textsc{( head entity, relation name, tail entity)}, e.g., \textsc{(Paris, IsCapitalOf, France)} \cite{hogan2021knowledge}.
Reasoning over KGs (such as predicting a missing link) has been a longstanding challenge in AI \cite{liang2024kgreasoningsurvey,nayyeri2021logicenn}. 
Recent structural knowledge graph foundation models (Structural KGFMs) \cite{DBLP:conf/iclr/0001YM0Z24,liu2025graphfm} seek to overcome the limitations of traditional transductive embedding methods \cite{nayyeri2021logicenn,wang2017knowledge} by enabling fully inductive generalization to unseen entities and relations. The core idea is to learn \textit{transferable structural patterns} instead of memorizing entities and relations.
Structural KGFMs build universal relational representations by constructing a relation graph (a graph where nodes are relations) \cite{lee2023ingram,DBLP:conf/iclr/0001YM0Z24} and applying message passing, thereby obtaining representations for new relations without any node or textual features.

While such foundation models have made important progress, they often inherit a key architectural limitation: the use of a single fixed relation transformation in the message-passing or scoring function. An element-wise multiplication (DistMult-style bilinear transform \cite{distmultyang2014embedding,DBLP:conf/iclr/0001YM0Z24}) is used to combine an entity with a relation when propagating messages.
Relying on a single, uniform transformation constrains the model’s capacity to represent the diverse relational patterns essential for structural knowledge graph foundation models trained across multiple heterogeneous graphs.
For instance, relation specific element wise multiplication in message passing has an antisymmetric nature and cannot properly model symmetric relations, nor can it adequately represent hierarchical orderings.
Relying on a single algebraic geometric transformation means the model is biased toward a particular class of relational structure, potentially leading to suboptimal generalization \cite{nayyeri20215}.

Geometric knowledge graph embeddings research \cite{nayyeri20215,pan2024hge} has shown that different geometric spaces offer complementary strengths. Complex number embeddings (as in ComplEx and RotatE) can model symmetry, anti-symmetry, and cyclic composition via rotations in the complex plane \cite{DBLP:conf/iclr/SunDNT19}. Hyperbolic or split-complex representations can naturally capture partial orders and hierarchical relations due to their ability to represent infinite or unbounded distances. Dual-number embeddings introduce translational components (via nilpotent $\epsilon$ terms) that can model one-to-many relations or additive offsets, 
while also enabling non-commutative compositions \cite{dong-etal-2024-dual,pan2024hge}.
Each of these algebraic families (in real, complex, split-complex, and dual spaces) provides a unique bias: no single space is optimal for all relation types. This raises a crucial question:

\begin{tcolorbox}[takeawaybox]
\textbf{Question:} Can we combine multiple geometric transformations to create a more powerful, universally generalizing Structural KGFM trained on multiple KGs with diverse relational patterns?
\end{tcolorbox}

\begin{figure*}[htbp]
    \centering
    \includegraphics[width=0.9\textwidth]{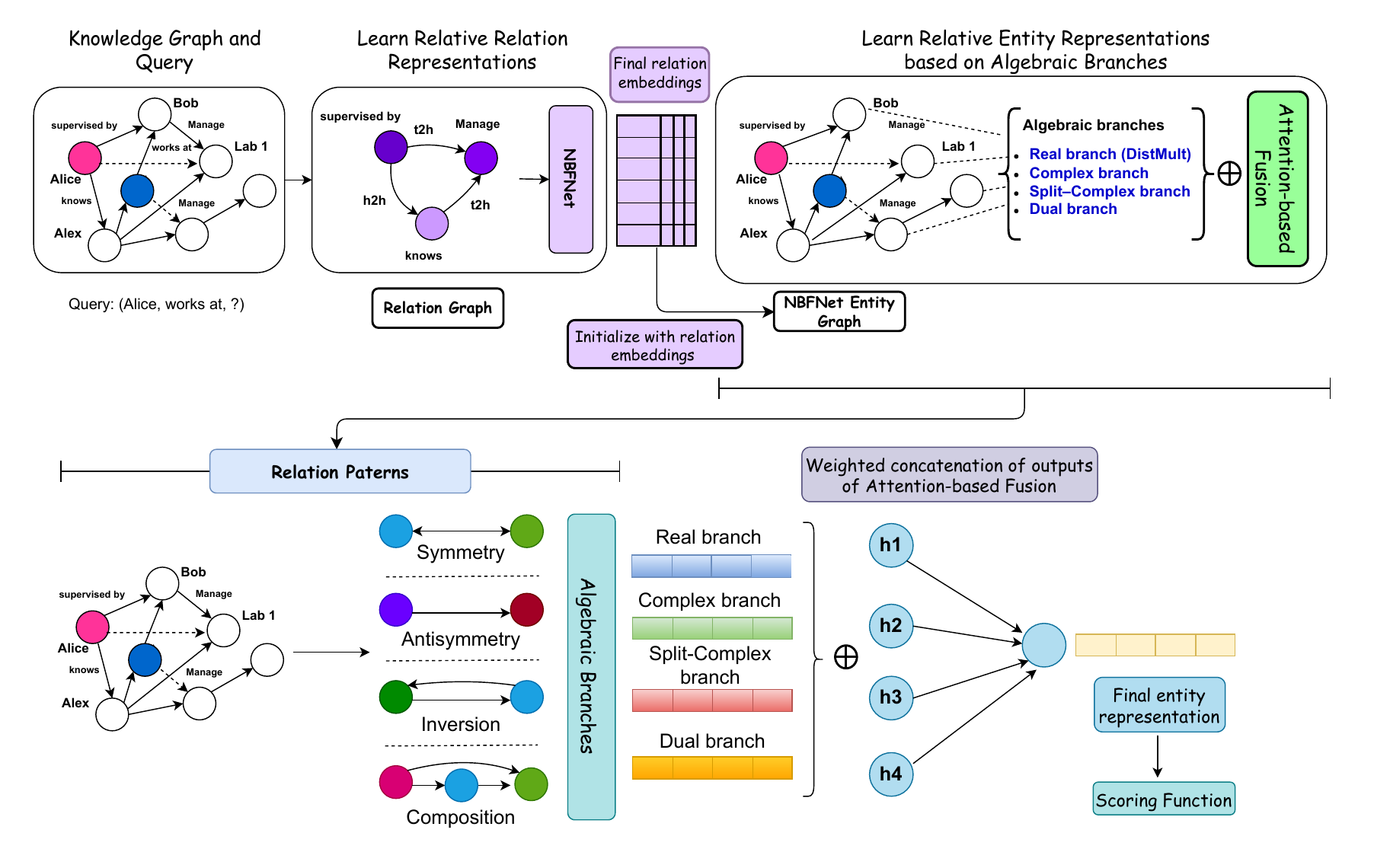}
    \caption{Overall architecture of the \textsc{Gamma} model.}
    \label{fig:architecture_gamma}
\end{figure*}

In this paper, we answer this question by proposing \textsc{Gamma} (Geometric Attention Multi-Message Aggregation), a novel structural knowledge graph foundation model that fuses multiple relational message-passing mechanisms. 
Instead of using one relation transform across the board, \textsc{Gamma} leverages multi-head message functions in parallel (Figure \ref{fig:architecture_gamma}), where each head operates in a different geometric space (real (flat), complex (sphere), split-complex (hyperbola), or dual (Galilean circle)). 
These heads produce diverse candidate messages for each triple, which are then aggregated by an attention module that learns to suitably combine them for the query at hand.
Intuitively, \textsc{Gamma} works as a stable and compact integrator to form complementary but not overly divergent features, which enables more expressive relational representations that lead to better generalization than any single space message-passing alone.
\IEEEpubidadjcol
We provide a full formalization of each message function and the attention-based fusion, and we derive insights into why this multi-head approach is more expressive.
In particular, we show that the combined representation can model important relational properties like symmetry, anti-symmetry, and composition more effectively than any single message function. From an architecture perspective, we carefully design \textsc{Gamma}’s fusion strategy to maintain separate “expert” branches for each message type through the layers, only merging their outputs before the final prediction stage, a choice that we empirically validate via ablation studies (demonstrating its superiority over naive layer-level fusion).

Empirically, we pre-train \textsc{Gamma} on the three source graphs (FB15k-237, WN18RR, and CoDEx-M \cite{toutanova2015observed,DBLP:conf/aaai/DettmersMS018,safavi-koutra-2020-codex}) and evaluate zero-shot on 53 unseen target graphs covering transductive, inductive entity, and inductive entity-relation scenarios.
\textsc{Gamma} consistently outperforms the \textsc{Ultra} \cite{DBLP:conf/iclr/0001YM0Z24} baseline (which uses only real element-wise multiplication message passing (DistMult-style bilinear transform)) in the link prediction task, with particularly large gains on the challenging inductive benchmarks.
Notably, we show that these improvements do not rely completely on increasing the model's depth or width (i.e., increasing the model capacity); rather, the performance boost comes from learning complementary relational biases. 
For example, on average \textsc{Gamma} improves MRR by 4.4\% and Hits@10 by 2.5\% across all 53 graphs compared to \textsc{Ultra}, with up to 7.0\% MRR on \textit{Inductive} $\textit{(e)}$ sets, while maintaining the same results on transductive sets (no loss of performance on traditional tasks).

Our contributions are summarized as follows:

\begin{itemize}
    \item \textit{Multi-Head Geometric Message Passing}: We introduce a new KG reasoning architecture that parallelizes multiple algebraic message-passing heads (real, complex, split-complex, dual), greatly enhancing expressiveness. To our knowledge, \textsc{Gamma} is the first foundation model to incorporate a mixture of geometric transformations for relational reasoning.

    \item \textit{Mathematical Formalism and Insight:}
    We provide a unified formal description of each algebraic head and prove how the combination can represent a strictly broader class of relational patterns than single-head models. We also discuss examples of relational structures that \textsc{Gamma} has a potentially stronger modeling capability from a theoretical perspective (e.g., one-to-many mappings with hierarchy and simultaneity), which are difficult for single-space models like DistMult or even ComplEx alone.

    \item \textit{Performance Improvement:} 
    Experimentally, 
    \textsc{Gamma} outperforms \textsc{Ultra} (the previous best foundation model) on 53 evaluation KGs on average, with particularly strong gains in the difficult inductive scenario. 
    Through rigorous ablations, we show that the necessity and effectiveness of stable multi-head attention gating: removing any component of the module leads to a degradation in the model’s capability; 
    simply increasing \textsc{Ultra}’s hidden dimension and dimensions of feed-forward network to match \textsc{Gamma}’s parameter count does not replicate our gains. 
    We also find that the complex + split-complex and complex + DistMult combinations yield the best synergy among message types, aligning with our message function complementarity hypothesis.

    \item \textit{Robustness and Generality:} \textsc{Gamma}’s robust attention gating allows it to adapt to a variety of graphs without retraining and yield consistent improvements without explicit branch selection.
    Unlike strong routing, we apply regularization to prevent gate collapse, forming a stable and compact integrator: weakening the regularization causes the attention to shift to single-branch dominance and leads to performance degradation.
\end{itemize}

The remainder of this paper is structured as follows: Section~\ref{sec:related_work} reviews existing literature on Knowledge Graph embeddings, geometric message passing, and foundation models, positioning our work within the current research landscape.
Section~\ref{sec:preliminaries} provides essential background and formal definitions relevant to Knowledge Graphs and geometric algebra utilized in our model.
Section~\ref{sec:methodology} details the proposed \textsc{Gamma} model, elucidating its novel multi-head geometric message passing architecture and attention-based fusion mechanism.
Section~\ref{sec:experiments} presents our comprehensive experimental evaluation, including dataset descriptions, baseline comparisons, performance results on zero-shot link prediction, and ablation studies.
Finally, Section~\ref{sec:conclusion} summarizes our findings, discusses the implications of \textsc{Gamma}, and outlines promising directions for future work.

\section{Related Work}
\label{sec:related_work}

\subsection{Structural KG Foundation Models}
Traditional KG embedding models \cite{liang2024kgreasoningsurvey,nayyeri2021logicenn} operate mainly in transductive and semi-inductive settings, where all entities, or at least most entities, and all relations are known during training, and the task is to infer missing links among them.
Early translation models like TransE \cite{NIPS2013_1cecc7a7} are scalable but weak for one to many relations, while bilinear or complex embeddings such as DistMult \cite{DBLP:journals/corr/YangYHGD14a} and ComplEx \cite{DBLP:conf/icml/TrouillonWRGB16} improve expressiveness. Later designs like ConvE \cite{DBLP:conf/aaai/DettmersMS018} enhances feature interaction with convolution, and GNNs like RGCN \cite{DBLP:conf/esws/SchlichtkrullKB18} and CompGCN \cite{DBLP:conf/iclr/VashishthSNT20} capture structural context through message passing yet remain tied to fixed embeddings. To address these limits, semi-inductive methods like GraIL \cite{DBLP:conf/icml/TeruDH20} employs enclosing subgraphs to generalize to unseen entities, and \textsc{NBFNet} \cite{DBLP:conf/nips/ZhuZXT21} integrates path reasoning with GNNs for stronger interpretability. A*Net \cite{DBLP:conf/nips/ZhuYGX0G023} further improves efficiency, while RED-GNN \cite{DBLP:conf/www/ZhangY22} utilizes relational digraphs for richer structure. AdaProp \cite{DBLP:conf/kdd/ZhangZY0023} adaptively samples paths to reduce noise, and NodePiece \cite{DBLP:conf/iclr/0001DWH22} tokenizes entities via anchors, cutting dependence on large embedding tables.

However, semi-inductive methods still rely on fixed relation vocabularies, limiting their generalization. This motivates the development of structural KG foundation models. These models extend the move to a fully inductive setting, where models generalize to unseen entities and relations across new graphs. Rather than memorizing embeddings, they exploit structural patterns, relation graphs, prompts, or motifs. INGRAM \cite{DBLP:conf/icml/LeeCW23} pioneered the study of fully inductive reasoning by building weighted relation graphs for unseen entities and relations, while RMPI \cite{DBLP:conf/icde/GengCPCJZC23} broadened this direction through local subgraph extraction at significant computational expense. \textsc{Ultra} \cite{DBLP:conf/iclr/0001YM0Z24} advanced the paradigm by learning universal structural motifs, inspiring successors such as TRIX \cite{DBLP:journals/corr/abs-2502-19512} and MOTIF \cite{DBLP:journals/corr/abs-2502-13339} to pursue more expressive motif-based reasoning, and GraphOracle \cite{DBLP:journals/corr/abs-2505-11125} introduced relation-dependency graphs to enhance cross-relation generalization. At the same time, ISDEA \cite{zhou2025doubleequivarianceinductivelink} and MTDEA \cite{zhou2023multitaskperspectivelinkprediction} established double equivariant formulations to guarantee invariance across nodes and relations, and KG-ICL \cite{DBLP:conf/nips/CuiSH24} later demonstrated that in-context learning with subgraph prompts could scale inductive reasoning across diverse knowledge graphs.

\subsection{Geometric and Algebraic KG Learning}
Geometric relational transformation-based methods frame knowledge graph reasoning as learning geometric operations in embedding space. Translation-based methods like TransE \cite{NIPS2013_1cecc7a7} view relations as vector shifts, offering scalability but limited expressiveness. Rotation-based approaches like RotatE \cite{DBLP:conf/iclr/SunDNT19} extend this idea into the complex plane, enabling modeling of symmetry, anti-symmetry, and relation composition. Hypercomplex embeddings like ComplEx \cite{DBLP:conf/icml/TrouillonWRGB16} and QuatE \cite{DBLP:conf/nips/0007TYL19} enrich representation power through complex and quaternion spaces, respectively. Non-Euclidean methods such as MuRP \cite{DBLP:conf/nips/BalazevicAH19} exploit hyperbolic geometry to capture hierarchical patterns, while region-based approaches like BoxE \cite{DBLP:conf/nips/AbboudCLS20} encode relations as hyper-rectangles that naturally capture inclusion and intersection. Compared to entity-focused or inductive foundation models, geometric methods emphasize algebraic and spatial transformations, offering strong expressiveness in capturing relation semantics but often struggling with full inductive generalization. These methods provide inductive biases for modeling relational patterns, which structural KG foundation models build upon to achieve fully inductive generalization beyond fixed embeddings.

\section{Preliminaries}
\label{sec:preliminaries}
This section introduces the fundamental concepts and notation necessary for defining our proposed model, \textsc{Gamma}, focusing on the structure of the knowledge graph foundation model and the task of inductive link prediction.

\subsection{Knowledge Graphs and Notation}
A Knowledge Graph (KG) is formally defined as 
\[
G = (E, R, T),
\] 
where $E$ is the set of entities (nodes), $R$ is the set of relations (edge types), and $T$ is the set of factual triples.  
A triple $(h, r, t) \in T$ represents a fact, where $h \in E$ is the head entity, $t \in E$ is the tail entity, and $r \in R$ is the relation connecting them.

\subsection{Inductive Link Prediction}
The primary task addressed in this work is \textit{Inductive Link Prediction}. Unlike the transductive setting, where all entities are known during training, the inductive setting requires the model to generalize knowledge learned from a source graph, $G_{\text{src}}$ to an entirely novel, unseen target graph, 
$G_{\text{tgt}}$. Specifically, $G_{\text{src}}$ and $G_{\text{tgt}}$ are disjoint in terms of their entity and relation sets:
\[
E_{\text{src}} \cap E_{\text{tgt}} = \emptyset, \,\,\,
R_{\text{src}} \cap R_{\text{tgt}} = \emptyset.
\]

The model, trained on $G_{\text{src}}$, must predict missing links (e.g., $(h,r,?)$ or $(?,r,t)$) within the new $G_{\text{tgt}}$. This scenario is crucial for assessing the transferability and universal generalization ability of foundation models.

\subsection{Relational and Structural Patterns}
Following the formulation in \cite{DBLP:conf/iclr/SunDNT19}, the formal definitions of several relational patterns used in knowledge graph analysis, including symmetry, anti-symmetry, inversion, and composition, are as follows:

\begin{itemize}
    \item \textbf{Symmetric Relation:}
A relation \(r \in R\) is said to be \emph{symmetric} when 
\[
\forall h,t \in E: (h,r,t) \in T \;\Longrightarrow\; (t,r,h) \in T.
\]
That is, whenever a triple holds in one direction, the reversed triple must
also appear in the graph.

    \item \textbf{Anti-Symmetric Relation:}
A relation \(r \in R\) is \emph{anti-symmetric} if 
\[
\forall h,t \in E: (h,r,t) \in T \;\Longrightarrow\; (t,r,h) \notin T.
\]
In other words, the validity of a triple precludes the existence of its
reverse counterpart.

    \item \textbf{Inverse Relation:}
A relation \(r \in R\) is an \emph{inverse} of another relation 
\(r_{\mathrm{inv}} \in R\) if 
\[
\forall h,t \in E: (h,r,t) \in T \;\Longrightarrow\; (t,r_{\mathrm{inv}},h) \in T.
\]
Moreover, if there exists some \(r' \in R\) with \(r' \neq r\) such that 
\(r'\) satisfies the above inverse property with respect to \(r\), then \(r\) is
classified as an inverse relation.

    \item \textbf{Composite Relation:}
A relation \(r \in R\) is considered a \emph{composition} of two
relations \(r_{1}, r_{2} \in R\) if
\begin{equation*}
    \begin{split}
        &\forall a,b,c \in E: (a,r_{1},b) \in T \;\wedge\; (b,r_{2},c) \in T \; \\ 
        &\Longrightarrow\; (a,r,c) \in T.
    \end{split}
\end{equation*}

That is, if a path of length two through \(r_{1}\) and \(r_{2}\) exists, then a
direct triple following relation \(r\) must also be present. Whenever this
condition holds, we refer to \(r\) as a composite relation.
\end{itemize}

\subsection{Algebraic Message Passing}
Modern structural KG foundation models often utilize a message passing paradigm to aggregate information from local neighborhoods and compute updated entity representations. In this context, the information between a head entity $h$ and a tail entity $t$ through relation $r$ is often modeled via a transformation function $f_r(h)$:
\[
\mathbf{t} \approx \textsc{MSG}(f_r(\mathbf{h})).
\]

A large class of effective models, including the baseline \textsc{Ultra}, relies on a single algebraic geometric transformation to define $f_r(h)$. Common examples of these single geometric operations include:

\begin{itemize}
    \item \textbf{Translation:} 
    \[
    f_r(h) = h + r,
    \]
    which models relations as simple translation vectors in $\mathbb{R}^d$ (e.g., TransE). This is primarily effective for modeling compositional and hierarchical patterns.
    
    \item \textbf{Rotation:} 
    \[
    f_r(h) = h \circ r,
    \]
    where $\circ$ represents element-wise multiplication in the complex or quaternion space, thus modeling a rotation in the embedding space (e.g., RotatE \cite{DBLP:conf/iclr/SunDNT19}, QuatE \cite{DBLP:conf/nips/0007TYL19}). This is effective for modeling cyclic and asymmetric relations.
    
    \item \textbf{Reflection/Projection:}  
    This class involves operations like matrix multiplication,
    \[
    f_r(h) = W_r h,
    \]
    or specific parameter sharing mechanisms that effectively model reflections or projections in the embedding space. This is essential for capturing symmetric and inversion properties (e.g., RESCAL \cite{DBLP:conf/icml/NickelTK11}, SimplE \cite{DBLP:conf/nips/Kazemi018}).
\end{itemize}

While these single-geometric approaches are expressive for specific relational properties, they introduce inherent biases that limit their ability to universally capture diverse relational structures, which is the primary challenge that \textsc{Gamma} aims to overcome.

\section{Methodology}
\label{sec:methodology}

This section presents a formal description of the proposed \emph{Geometric Attention Multi\-Message Aggregation} (\textsc{Gamma}) model. As illustrated in Figure \ref{fig:architecture_gamma}, at a high level, \textsc{Gamma} extends the 
Neural Bellman–Ford network (\textsc{NBFNet}) framework \cite{DBLP:conf/nips/ZhuZXT21,DBLP:conf/iclr/0001YM0Z24} by 
(i) learning a representation for every relation by propagating signals over a 
\emph{relation graph} and 
(ii) employing several algebraically distinct message functions to propagate information over the 
\emph{entity graph}.  A trainable attention module then fuses the outputs of these message functions to produce a 
single representation conditioned on the query.

\subsection{Relation–Graph Learning}
Let $G=(E,R,T)$ be an input knowledge graph.
We define an auxiliary \emph{relation graph}
\begin{equation}
  G_r=(R,\mathcal{E}_r),
\end{equation}
whose node set coincides with the relation set~$R$.  The edge set $\mathcal{E}_r$ captures how relations co\-occur in the knowledge graph.
In order to distinguish different types of co\-occurrence, four directed edge types are introduced as follows:
\begin{itemize}
  \item \textbf{Head–to–Head (\textnormal{H2H}).}  There is a directed edge $r_i\xrightarrow{\text{H2H}}r_j$ in $\mathcal{E}_r$ if there exist entities $h\in E$ and $t_1,t_2\in E$ such that both $(h,r_i,t_1)$ and $(h,r_j,t_2)$ belong to~$T$.  This edge type indicates that $r_i$ and $r_j$ share a head entity.
  \item \textbf{Head–to–Tail (\textnormal{H2T}).}  There is an edge $r_i\xrightarrow{\text{H2T}}r_j$ if there exist triples $(h,r_i,m)$ and $(m,r_j,t)$ in~$T$.  In other words, the tail entity of $r_i$ coincides with the head entity of $r_j$, so that $r_j$ may follow $r_i$ along a path from a head entity to a tail entity.
  \item \textbf{Tail–to–Head (\textnormal{T2H}).}  An edge is introduced $r_i\xrightarrow{\text{T2H}}r_j$ if there exist triples $(m,r_i,h)$ and $(t,r_j,m)$ in~$T$.  This type links relations whose heads and tails are connected in the reverse order of \textnormal{H2T}.
  \item \textbf{Tail–to–Tail (\textnormal{T2T}).}  Finally, an edge  $r_i\xrightarrow{\text{T2T}}r_j$ is added when there exist triples $(h_1,r_i,t)$ and $(h_2,r_j,t)$ in~$T$.  In this case $r_i$ and $r_j$ share a tail entity.
\end{itemize}
Each edge $\eta=(r_i,\tau,r_j)\in\mathcal{E}_r$ therefore has a type $\tau\in\{\text{H2H},\text{H2T}$
$,\text{T2H},\text{T2T}\}$.  

To obtain a vector representation for every relation $r\in R$,
we apply \textsc{NBFNet} \cite{DBLP:conf/nips/ZhuZXT21,DBLP:conf/iclr/0001YM0Z24} to the relation graph $G_r$.
We denote by $\mathbf{h}^{(t)}_{r}$ the vector representation of the relation~$r$ at iteration~$t$. We initialize these representations with $\mathbf{1}_d$  for $r_q$, and $\mathbf{0}_d$  for the rest of the relations. 
For each iteration $t\geq 1$ the update rule reads
\begin{equation}
  \label{eq:relation-nbfnet}
  \mathbf{h}^{(t)}_{r_j} = \mathrm{AGG}_{\eta=(r_i,\tau,r_j)\in\mathcal{E}_r} \Bigl( \mathrm{MSG}(\mathbf{h}^{(t-1)}_{r_i}, \mathbf{e}_\tau) \Bigr),
\end{equation}
where $\mathrm{MSG}$ is a neural message function and $\mathrm{AGG}$ is a permutation\-invariant aggregator (e.g., a learnable sum or mean). 
Each edge type $\tau$ is associated with a trainable type embedding $\mathbf{e}_\tau\in\mathbb{R}^d$.  After $L$ iterations, we set the relation representation $\mathbf{r}=\mathbf{h}^{(L)}_r$.  
These relation embeddings are subsequently used to modulate the message passing over the entity graph.

\subsection{Entity–Graph Learning}
Given a query triple $q=(h,r_q, ?)$ consisting of a head entity $h$ and a query relation $r_q$, \textsc{Gamma} computes a representation of each candidate tail entity $e$ conditioned on $q$.  We achieve this by running several \emph{algebraically distinct} message passing processes on the entity graph and then aggregating their outputs with attention.

\paragraph{Generalized relation transformations.}
Let's $\theta$ denote the parameters of a particular message branch.  Each branch defines a relation-specific transformation
\begin{equation}
  f_{r}^{(\theta)} : \mathbb{K}^d \to \mathbb{K}^d,
\end{equation}
where $\mathbb{K}$ is an algebraic number system (e.g., the reals, complex numbers, split-complex numbers, or dual numbers).  Given an entity embedding $\mathbf{x}\in\mathbb{K}^d$ and a relation embedding $\mathbf{r}\in\mathbb{K}^d$, the transformed message $f_{r}^{(\theta)}(\mathbf{x})$ is computed by a fixed algebraic operation in $\mathbb{K}$.  
For the branches considered in this work, we obtain the following specific forms:
\begin{itemize}
  \item \textbf{Real (DistMult) branch.}  Working over $\mathbb{K}=\mathbb{R}$, the relation transformation is defined elementwise as
  \begin{equation}
    f_{r}^{\mathrm{real}}(\mathbf{x})\;=\;\mathbf{x}\odot\mathbf{r},
  \end{equation}
  where $\odot$ denotes the Hadamard (elementwise) product.  This recovers the bilinear DistMult operator.

  \item \textbf{Complex branch.}  Over the field of complex numbers $\mathbb{K}=\mathbb{C}$, each vector is represented by its real and imaginary parts, $\mathbf{x}=(\mathbf{x}_{\mathrm{re}},\mathbf{x}_{\mathrm{im}})$ and $\mathbf{r}=(\mathbf{r}_{\mathrm{re}},\mathbf{r}_{\mathrm{im}})$.  Complex multiplication yields
  \begin{equation}
  \begin{split}  
    &f_{r}^{\mathrm{complex}}(\mathbf{x})\;\\
    &=\;\bigl(\mathbf{x}_{\mathrm{re}}\odot\mathbf{r}_{\mathrm{re}} - \mathbf{x}_{\mathrm{im}}\odot\mathbf{r}_{\mathrm{im}},\;\mathbf{x}_{\mathrm{re}}\odot\mathbf{r}_{\mathrm{im}} + \mathbf{x}_{\mathrm{im}}\odot\mathbf{r}_{\mathrm{re}}\bigr).
    \end{split}
  \end{equation}
  This branch, therefore, implements rotations in the complex plane and is well suited to modelling symmetric and anti-symmetric relations.

  \item \textbf{Split–complex branch.}  The split–complex numbers introduce an imaginary unit $j$ satisfying $j^2=+1$.  For vectors $\mathbf{x}=(\mathbf{x}_{\mathrm{re}},\mathbf{x}_{\mathrm{im}})$ and $\mathbf{r}=(\mathbf{r}_{\mathrm{re}},\mathbf{r}_{\mathrm{im}})$ we define the split–complex multiplication as
  \begin{equation}
  \begin{split}  
    &f_{r}^{\mathrm{split}}(\mathbf{x})\;\\
    &=\;\bigl(\mathbf{x}_{\mathrm{re}}\odot\mathbf{r}_{\mathrm{re}} + \mathbf{x}_{\mathrm{im}}\odot\mathbf{r}_{\mathrm{im}},\;\mathbf{x}_{\mathrm{re}}\odot\mathbf{r}_{\mathrm{im}} + \mathbf{x}_{\mathrm{im}}\odot\mathbf{r}_{\mathrm{re}}\bigr).
    \end{split}
  \end{equation}
  Because $j^2=+1$, this operator can model hyperbolic rotations and thus enhances the ability to capture hierarchical and partial order patterns.

  \item \textbf{Dual branch.}  Dual numbers take the form $a+\varepsilon b$ with $\varepsilon^2=0$.  Writing $\mathbf{x}=(\mathbf{x}_{\mathrm{re}},\mathbf{x}_{\mathrm{im}})$ and $\mathbf{r}=(\mathbf{r}_{\mathrm{re}},\mathbf{r}_{\mathrm{im}})$, multiplication is given by
  \begin{equation}
    f_{r}^{\mathrm{dual}}(\mathbf{x})\;=\;\bigl(\mathbf{x}_{\mathrm{re}}\odot\mathbf{r}_{\mathrm{re}},\;\mathbf{x}_{\mathrm{re}}\odot\mathbf{r}_{\mathrm{im}} + \mathbf{x}_{\mathrm{im}}\odot\mathbf{r}_{\mathrm{re}}\bigr).
  \end{equation}
  The nilpotent nature of $\varepsilon$ allows this branch to encode translational offsets and one-to-many relations.
\end{itemize}

\paragraph{Conditional message passing.}
Fix a branch $k$ and its associated transformation $f^{(k)}$.  To compute a representation of entities conditioned on the source entity $h$ and query relation $r_q$, we run $T$ steps of a Bellman-Ford-style message passing on the entity graph $G$.
Let $\mathbf{z}_{u}^{(k,0)} \in \mathbb{K}^d$ denote the initial representation of each entity $u\in E$ for branch~$k$.  
We set $\mathbf{z}_{h}^{(k,0)}=\mathbf{1} * \mathbf{r}_q$ ($\mathbf{1}$ is a vector of ones) and $\mathbf{z}_{u}^{(k,0)}=\mathbf{0}$ for $u\neq h$.  At iteration $t\geq 1$, each entity $v$ aggregates messages from its incoming neighbors
\begin{equation}
  \label{eq:entity-message}
  \mathbf{m}_{u\to v}^{(k,t)} \;=\; f_{r}^{(k)}\bigl(\mathbf{z}_{u}^{(k,t-1)}\bigr), \quad \text{for every triple } (u,r,v) \in T,
\end{equation}
and updates its representation via a permutation-invariant aggregator $\mathrm{AGG}$,
\begin{equation}
  \label{eq:entity-aggregate}
  \mathbf{z}_{v}^{(k,t)} \;=\; \mathrm{AGG}\bigl(\{\mathbf{m}_{u\to v}^{(k,t)} : (u,r,v)\in T\}\bigr).
\end{equation}
This iterative process implicitly sums over all paths of length up to~$T$ starting from the source entity~$h$.  After $T$ iterations, we obtain a branch-specific representation $\mathbf{z}_{e}^{(k,T)}$ for every entity $e\in E$.

\paragraph{Attention–based fusion.}
The final step of \textsc{Gamma} combines the $K$ branch outputs using an attention mechanism conditioned on the query.  
First, a linear map $\mathbf{W}_\mathrm{ctx}$ projects the query relation in $q=(h,r_q)$ to a context vector $\mathbf{c}\in\mathbb{R}^{d_{\mathrm{att}}}$:
\begin{equation}
  \mathbf{c}\;=\;\mathrm{Norm}(\mathbf{W}_\mathrm{ctx}\,[\mathbf{r}_q]),
\end{equation}
where $\mathrm{Norm}(\cdot)$ denotes $\mathrm{L}_2$-normalization and $\mathbf{r}_q$ is a learned embedding of the query relation
from~\eqref{eq:relation-nbfnet}.  
For each branch $k$, we similarly project the entity representation $\mathbf{z}_e^{(k,T)}$ into the same attention space via $\mathbf{W}_\mathrm{key}$ to obtain a key vector $\mathbf{k}_e^{(k)} = \mathrm{Norm}(\mathbf{W}_\mathrm{key}\,\mathbf{z}_e^{(k,T)})$.  The attention weight of branch $k$ for entity~$e$ is then
\begin{equation}
  \label{eq:attention-weight}
  \alpha_e^{(k)}\;=\;\frac{\exp(\cos(\mathbf{k}_e^{(k)},\; \mathbf{c}) / \kappa)}{\sum{k'}\exp(\cos(\mathbf{k}_e^{(k')},\; \mathbf{c}) / \kappa)},
\end{equation}
where $\kappa$ is a temperature parameter controlling the sharpness of the attention distribution.
To avoid over-focusing on a single branch, the final attention weights are mixed with a uniform distribution:
\begin{equation}
  \label{eq:uniform-mixing}
  \tilde{\alpha}_e^{(k)}\;=\;(1 - \lambda) \alpha_e^{(k)} + \lambda \cdot \frac{1}{k},
\end{equation}
where $\lambda \in [0,1]$ is a uniform mixing coefficient.
Finally, the entity representation conditioned on the query is a concatenation of rescaled branch outputs:
\begin{equation}
  \mathbf{z}_{e}\;=\;\text{Concat}\left(
  \tilde{\alpha}_e^{(1)}\odot\mathbf{z}_e^{(1,T)},\;\dots,\;
  \tilde{\alpha}_e^{(K)}\odot\mathbf{z}_e^{(K,T)}\right).
\end{equation}
To score a candidate tail entity $e$, we apply a feed-forward network $\psi:\mathbb{R}^d\to\mathbb{R}$ to the real part of $\mathbf{z}_e$ and compute
\begin{equation}
  \mathrm{score}(h,r_q,e)\;=\;\psi\bigl((\mathbf{z}_e)\bigr).
\end{equation}
During training, to encourage diversity among attention distributions and mitigate branch collapse, an entropy-based regularization term is incorporated:
\begin{equation}
  \mathcal{L}_{\text{ent}}\;=\;-\frac{1}{N}\sum_{e=1}^{N}\sum_{k=1}^{K}\tilde{\alpha}_e^{(k)}\log \tilde{\alpha}_e^{(k)}.
\end{equation}
The overall training objective is to minimize $\mathcal{L}$, the primary prediction loss $\mathcal{L}_{\text{pred}}$ (a negative log-likelihood over positive and negative triplets as described in \cite{DBLP:conf/nips/ZhuZXT21}) combined with the entropy regularizer:
\begin{equation}
  \mathcal{L}\;=\;\mathcal{L}_{\text{pred}} - \beta \mathcal{L}_{\text{ent}},
\end{equation}
where $\beta$ is a small coefficient controlling the strength of regularization.

The use of multiple algebraic branches endows \textsc{Gamma} with the ability to model a wide range of relational patterns.  Complex multiplication captures cyclic and anti-symmetric relations, split–complex multiplication models hierarchical and hyperbolic interactions, while dual multiplication accounts for translational offsets. The attention mechanism in~\eqref{eq:attention-weight} learns to weight these branches depending on the query, thereby enabling the model to adaptively balances the contributions of heterogeneous message passings, thereby enhancing expressiveness, interpretability, and generalization across complex relational graphs.

\section{Experiments}
\label{sec:experiments}
By conducting evaluations across diverse knowledge graphs, we seek to answer the following research questions: \textbf{RQ1}: To what extent does multi-head geometric attention enhance the generalization ability of \textsc{\textsc{Ultra}}? \textbf{RQ2}: What underlying benefits enable multi-head geometric attention to surpass a single message function? \textbf{RQ3}: Do the gains come from attention or simply from more parameters? \textbf{RQ4}: How do the fusion modules influence performance?

\subsection{Experiment Setup}
We build upon the open-source PyG (PyTorch Geometric) implementation of \textsc{Ultra} \cite{DBLP:conf/iclr/0001YM0Z24} by modifying the entity model architecture to introduce a multi-head geometric attention mechanism. We leave improvements to the relation model architecture for future work. 

Our experimental setup remains consistent with \textsc{Ultra}: Besides replacing the original single DistMult \cite{DBLP:journals/corr/YangYHGD14a} message function in the EntityModel with different combinations of message functions (selected from DistMult, complex \cite{DBLP:conf/icml/TrouillonWRGB16}, split-complex \cite{DBLP:conf/aaai/PanNLS24}, and dual \cite{dong-etal-2024-dual}), all other hyperparameters remain identical to \textsc{Ultra}. 
We provide more details in Appendix B in the supplementary material.

For pre-training, we use the same three datasets and 53 datasets for zero-shot evaluation. Due to the extremely large scale of Hetionet \cite{10.7554/eLife.26726} used in \textsc{Ultra} \cite{DBLP:conf/iclr/0001YM0Z24}, we exclude it from our evaluation. This dataset requires substantially higher computational resources and a much longer time than our current setting allows. Importantly, the remaining datasets cover a wide range of domains and scales, providing a representative and comprehensive evaluation of model generalization. During evaluation, the best checkpoint is selected based on its performance on validation sets: VGCS (Validation-Guided Checkpoint Strategy). In this setting, we use the 10 checkpoints saved after each epoch during pre-training to identify the one achieving the highest average MRR across 53 validation sets, and use it to report the final results on the 53 test sets. The resulting model, \textsc{Gamma}, contains approximately 359K parameters. Pre-training is carried out on four NVIDIA H200 GPUs, taking around 15 hours for 10 epochs. 
We provide the code in the supplementary material.

\subsection{Datasets}
Our experiments cover 56 publicly available knowledge graph datasets from diverse domains and sizes. These datasets are organized into three generalization scenarios: 
\begin{itemize}
    \item \textbf{15 \textit{Transductive} datasets} with fixed entities and relations across training and inference: WN18RR \cite{DBLP:conf/aaai/DettmersMS018}, FB15k-237 \cite{toutanova2015observed}, CoDEx-M \cite{safavi-koutra-2020-codex}, CoDEx-S \cite{safavi-koutra-2020-codex}, CoDEx-L \cite{safavi-koutra-2020-codex}, NELL-995 \cite{xiong-etal-2017-deeppath}, YAGO310 \cite{Mahdisoltani2015YAGO3AK}, WD-singer \cite{lv-etal-2020-dynamic}, NELL23K \cite{lv-etal-2020-dynamic}, FB15k-237-10\% \cite{lv-etal-2020-dynamic}, FB15k-237-20\% \cite{lv-etal-2020-dynamic}, FB15k-237-50\% \cite{lv-etal-2020-dynamic}, DB100K \cite{ding-etal-2018-improving}, Aristo-V4 \cite{DBLP:conf/akbc/ChenM0S21}, ConceptNet-100K \cite{DBLP:conf/aaai/MalaviyaBBC20}, among them, FB15k-237, WN18RR, and CoDEx-M are used for pre-training.
    \item \textbf{18 \textit{Inductive} $\textbf{\textit{(e)}}$ datasets} where new entities emerge at inference while relations remain fixed: WN18RR:v1 \cite{DBLP:conf/icml/TeruDH20}, WN18RR:v2 \cite{DBLP:conf/icml/TeruDH20}, WN18RR:v3 \cite{DBLP:conf/icml/TeruDH20}, WN18RR:v4 \cite{DBLP:conf/icml/TeruDH20}, FB15k-237:v1 \cite{DBLP:conf/icml/TeruDH20}, FB15k-237:v2 \cite{DBLP:conf/icml/TeruDH20}, FB15k-237:v3 \cite{DBLP:conf/icml/TeruDH20}, FB15k-237:v4 \cite{DBLP:conf/icml/TeruDH20}, NELL-995:v1 \cite{DBLP:conf/icml/TeruDH20}, NELL-995:v2 \cite{DBLP:conf/icml/TeruDH20}, NELL-995:v3 \cite{DBLP:conf/icml/TeruDH20}, NELL-995:v4 \cite{DBLP:conf/icml/TeruDH20}, ILPC22-S \cite{DBLP:journals/corr/abs-2203-01520}, ILPC22-L \cite{DBLP:journals/corr/abs-2203-01520}, Hamaguchi-BM:1k \cite{DBLP:conf/ijcai/HamaguchiOSM17}, Hamaguchi-BM:3k \cite{DBLP:conf/ijcai/HamaguchiOSM17}, Hamaguchi-BM:5k \cite{DBLP:conf/ijcai/HamaguchiOSM17}, INDIGO-BM \cite{NEURIPS2021_0fd600c9}.
    \item \textbf{23 \textit{Inductive} $\textbf{\textit{(e,r)}}$ datasets} where both new entities and relations emerge at inference: FB-100 \cite{pmlr-v202-lee23c}, FB-50 \cite{pmlr-v202-lee23c}, FB-75 \cite{pmlr-v202-lee23c}, FB-25 \cite{pmlr-v202-lee23c}, WK-100 \cite{pmlr-v202-lee23c}, WK-50 \cite{pmlr-v202-lee23c}, WK-75 \cite{pmlr-v202-lee23c}, WK-25 \cite{pmlr-v202-lee23c}, NL-100 \cite{pmlr-v202-lee23c}, NL-75 \cite{pmlr-v202-lee23c}, NL-50 \cite{pmlr-v202-lee23c}, NL-25 \cite{pmlr-v202-lee23c}, NL-0 \cite{pmlr-v202-lee23c}, \textsc{WikiTopics-MT1:Tax} \cite{DBLP:journals/corr/abs-2307-06046}, \textsc{WikiTopics-MT1:Health} \cite{DBLP:journals/corr/abs-2307-06046}, \textsc{WikiTopics-MT2:Org} \cite{DBLP:journals/corr/abs-2307-06046}, \textsc{WikiTopics-MT2:Sci} \cite{DBLP:journals/corr/abs-2307-06046}, \textsc{WikiTopics-MT3:Art} \cite{DBLP:journals/corr/abs-2307-06046}, \textsc{WikiTopics-MT3:Infra} \cite{DBLP:journals/corr/abs-2307-06046}, \textsc{WikiTopics-MT4:Sci} \cite{DBLP:journals/corr/abs-2307-06046}, \textsc{WikiTopics-MT4:Health} \cite{DBLP:journals/corr/abs-2307-06046}, \textsc{MetaFam} \cite{DBLP:journals/corr/abs-2307-06046}, FBNELL \cite{DBLP:journals/corr/abs-2307-06046}.
\end{itemize}
We provide the full description of these datasets in Appendix A in the supplementary material.

\subsection{Task}
During evaluation, we apply \textsc{Gamma} in a zero-shot setting, meaning the model is not trained or fine-tuned on the target datasets. The link prediction task involves predicting missing head or tail entities; however, consistent with \textsc{Ultra}, only tail prediction is conducted for the three datasets (FB15k-237-10\%, FB15k-237-20\%, FB15k-237-50\%) introduced by \cite{lv-etal-2020-dynamic}. We adopt the filtered ranking protocol \cite{DBLP:conf/nips/BordesUGWY13} and report Mean Reciprocal Rank (MRR) along with Hits@10 (H@10) as the primary evaluation metrics, computed against the entire set of entities in the inference graph and under the VGCS setting. According to \cite{DBLP:conf/iclr/0001YM0Z24}, zero-shot inference produces deterministic results; each evaluation is executed once. 

\subsection{Baselines}
To highlight the benefits of multi-head geometric attention over using a single message function, we take \textsc{Ultra} \cite{DBLP:conf/iclr/0001YM0Z24} as our baseline model for comparison. All baseline results are reproduced using \textsc{Ultra}’s official PyG implementation. While the results reported in \textsc{Ultra}'s original paper were based on the TorchDrug framework, we adopt the PyG framework as its tensor-based message passing paradigm and customizable operator interfaces align well with our design of multi-branch stacking and attention fusion in \textsc{NBFNet} \cite{DBLP:conf/nips/ZhuZXT21}. Compared to TorchDrug’s higher-level graph abstraction, PyG offers more flexible low-level control and more efficient sparse computation support, which facilitates architectural extensions and performance optimization. 

\subsection{To What Extent Does Multi-head Geometric Attention Enhance the Generalization Ability of \textsc{\textit{Ultra}}?}

\begin{table*}[htbp]
\caption{Zero-Shot Link Prediction Average Results of \textsc{Gamma} and \textsc{Ultra} Grouped by Datasets Generalization Scenarios over 53 KGs.}
\label{tab:average_results}
\centering
\resizebox{0.9\textwidth}{!}{
\begin{tabular}{llcccccccc}
\toprule
\multirow{2}{*}{\textbf{Model}} & \multirow{2}{*}{\textbf{Message Type}} &
\multicolumn{2}{c}{\makecell{\textbf{Inductive $e,r$} \\ (23 graphs)}} &
\multicolumn{2}{c}{\makecell{\textbf{Inductive $e$} \\ (18 graphs)}} &
\multicolumn{2}{c}{\makecell{\textbf{Transductive} \\ (12 graphs)}} &
\multicolumn{2}{c}{\makecell{\textbf{Total Avg} \\ (53 graphs)}} \\
\cmidrule(lr){3-4} \cmidrule(lr){5-6} \cmidrule(lr){7-8} \cmidrule(lr){9-10}
& & \textbf{MRR} & \textbf{H@10}
  & \textbf{MRR} & \textbf{H@10}
  & \textbf{MRR} & \textbf{H@10}
  & \textbf{MRR} & \textbf{H@10} \\
\midrule
\textsc{\textsc{Ultra}} & DistMult & 0.346 & 0.511 & 0.412 & 0.560 & \textbf{0.310} & 0.446 & 0.360 & 0.513 \\
\textsc{\textsc{Gamma}} & complex \& split-complex & \textbf{0.360} & \textbf{0.519} & \textbf{0.441} & \textbf{0.578} & \textbf{0.310} & \textbf{0.460} & \textbf{0.376} & \textbf{0.526} \\
\bottomrule
\end{tabular}
}
\end{table*}

Table \ref{tab:average_results} shows the average zero-shot results of \textsc{Gamma} and \textsc{Ultra} \cite{DBLP:conf/iclr/0001YM0Z24} on 53 graphs. \textsc{Gamma} consistently outperforms baselines on inductive datasets while maintaining the same performance on transductive datasets. Its advantage is most pronounced in the inductive setting; this can be attributed to the smaller graph scale of inductive datasets. Here, \textsc{Gamma} achieves improvements of 7\% in average MRR and 3.2\% in average Hits@10 on \textit{Inductive} $\textit{(e)}$ benchmarks, highlighting the effectiveness of multi-head geometric attention to compensate for the representational limitations of a single message function like DistMult \cite{DBLP:journals/corr/YangYHGD14a}. Even when averaged across all 53 datasets, \textsc{Gamma} still yields a consistent boost (4.4\% in MRR, 2.5\% in Hits@10), suggesting that its improvements are not isolated to a few graphs but robust across scales and domains. 
We provide detailed results for each dataset in Appendix D in the supplementary material.

In the field of knowledge graph link prediction, improvements such as those reported by Low-Dimensional Hyperbolic KG Embeddings \cite{DBLP:conf/acl/ChamiWJSRR20} (6.1\% MRR improvement) are already regarded as significant. This provides a basis for considering the improvements in the generalization ability of \textsc{Ultra} in knowledge graph reasoning achieved by our model as statistically and practically significant.

\subsection{What Underlying Benefits Enable Multi-head Geometric Attention to Surpass a Single Message Function?}
Across the 53 evaluation datasets, \textsc{Gamma} outperforms \textsc{Ultra} \cite{DBLP:conf/iclr/0001YM0Z24} on 47 datasets in terms of MRR and on 44 datasets in Hits@10. The improvements are particularly pronounced on smaller inductive benchmarks. For example, in the \textit{Inductive} $\textit{(e,r)}$ setting, NL-75 \cite{pmlr-v202-lee23c} and \textsc{WikiTopics-MT2:Sci} \cite{DBLP:journals/corr/abs-2307-06046} show MRR gains of 16.2\% and 14.2\%, respectively. Similarly, in the \textit{Inductive} $\textit{(e,r)}$ setting, WN18RR:v1 \cite{DBLP:conf/icml/TeruDH20}, Hamaguchi-BM:3k \cite{DBLP:conf/ijcai/HamaguchiOSM17}, and Hamaguchi-BM:5k \cite{DBLP:conf/ijcai/HamaguchiOSM17} exhibit substantial improvements of 62.0\%, 20.9\%, and 17.9\% in MRR, respectively.

We relate this to the high prevalence of symmetric, strongly anti-symmetric, or strongly composition relations in these datasets. In such settings, the limitations of DistMult \cite{DBLP:journals/corr/YangYHGD14a} are exposed. By contrast, the nature of complex \cite{DBLP:conf/icml/TrouillonWRGB16} and split-complex numbers \cite{DBLP:conf/aaai/PanNLS24} are more capable in this aspect. 
Under our conditional expert selection mechanism, the complementarity between complex and split-complex, in terms of their functional inductive biases and gradient diversity, should be effectively utilized. Hence, we hypothesize that such complementary branches may contribute to stronger zero-shot generalization.

To verify this hypothesis, we follow the methodology from \cite{DBLP:journals/pami/AliBHVGSFTL22} to identify symmetric, anti-symmetric, and compositional relational patterns across all test datasets. 
Based on these detected patterns, we partition the test triples into three corresponding subsets for evaluation. Since not all test sets contain enough triples of all three relational patterns, we only evaluate and report results for a relational pattern group when it includes at least 50 test triples to ensure statistical reliability. 
We provide the detailed subset statistics in Appendix A in the supplementary material.

We present the zero-shot average MRR and Hits@10 of \textsc{Ultra} \cite{DBLP:conf/iclr/0001YM0Z24} and \textsc{Gamma} on three relational pattern subsets derived from 53 test sets in Table \ref{tab:pattern-subsets}. On average, \textsc{Gamma} outperforms \textsc{Ultra} on all three relational pattern subsets.

\begin{tcolorbox}[takeawaybox]
\textbf{Takeaway 1.} The complementary inductive biases encoded by different branches allow the multi-head geometric attention mechanism to more effectively extract useful relational cues, thereby enhancing the model’s expressive capacity.
\end{tcolorbox}

\begin{table}[htbp]
\caption{Zero-Shot Link Prediction Average Results of \textsc{Ultra} and \textsc{Gamma} Grouped by Three Relational Patterns.}
\label{tab:pattern-subsets}
\centering
\setlength{\tabcolsep}{6pt}
\renewcommand{\arraystretch}{1.15}
\begin{tabular}{lcc cc cc}
\toprule
\multirow{2}{*}{\textbf{Model}} &
\multicolumn{2}{c}{\makecell{\textbf{\textbf{symmetric}} \\ (12 graphs)}} &
\multicolumn{2}{c}{\makecell{\textbf{\textbf{anti-symmetric}} \\ (53 graphs)}} &
\multicolumn{2}{c}{\makecell{\textbf{\textbf{composition}} \\ (39 graphs)}} \\
\cmidrule(lr){2-3}\cmidrule(lr){4-5}\cmidrule(lr){6-7}
& \textbf{MRR} & \textbf{H@10}
& \textbf{MRR} & \textbf{H@10}
& \textbf{MRR} & \textbf{H@10} \\
\midrule
\textsc{Ultra} & 0.824 & 0.871 & 0.316 & 0.468 & 0.355 & 0.515 \\
\textsc{Gamma} & \textbf{0.854} & \textbf{0.879} & \textbf{0.324} & \textbf{0.477} & \textbf{0.363} & \textbf{0.526} \\
\bottomrule
\end{tabular}
\end{table}

\subsection{Do the Gains Come from Attention or Simply from More Parameters?}
To isolate the effect of model capacity from the contribution of multi-head geometric attention, we trained two additional variants of \textsc{Ultra} \cite{DBLP:conf/iclr/0001YM0Z24}. One variant increases the hidden dimension and MLP width to approximately match GAMMA’s parameter count, while another variant directly uses two parallel DistMult branches of identical structure. These two expansion strategies allow us to tease apart the improvements contributed by the attention mechanism and those stemming from the enhanced geometric expressiveness of the message functions.

\begin{table*}[htbp]
\caption{Zero-Shot Link Prediction Average Results of \textsc{Gamma}, \textsc{Ultra Cap+} and \textsc{Ultra Att} Grouped by Datasets Generalization Scenarios over 53 KGs. \textsc{Ultra Cap+} Increases the Hidden Dimension and MLP Width to Approximately Match GAMMA’s Parameter Count, while \textsc{Ultra Att} Uses Two Parallel DistMult Branches of Identical Structure Under Our Multi-Head Attention Mechanism.}
\label{tab:parameter_comparison}
\centering
\resizebox{0.9\textwidth}{!}{
\begin{tabular}{llcccccccc}
\toprule
\multirow{2}{*}{\textbf{Model}} & \multirow{2}{*}{\textbf{Parameter Size}} &
\multicolumn{2}{c}{\makecell{\textbf{Inductive $e,r$} \\ (23 graphs)}} &
\multicolumn{2}{c}{\makecell{\textbf{Inductive $e$} \\ (18 graphs)}} &
\multicolumn{2}{c}{\makecell{\textbf{Transductive} \\ (12 graphs)}} &
\multicolumn{2}{c}{\makecell{\textbf{Total Avg} \\ (53 graphs)}} \\
\cmidrule(lr){3-4} \cmidrule(lr){5-6} \cmidrule(lr){7-8} \cmidrule(lr){9-10}
& & \textbf{MRR} & \textbf{H@10}
  & \textbf{MRR} & \textbf{H@10}
  & \textbf{MRR} & \textbf{H@10}
  & \textbf{MRR} & \textbf{H@10} \\
\midrule
\textsc{Ultra Cap+} & 360,395 & 0.344 & \textbf{0.524} & 0.435 & 0.574 & 0.302 & 0.446 & 0.365 & 0.523 \\
\textsc{Ultra Att} & 359,298 & 0.349 & \textbf{0.524} & 0.438 & 0.573 & \textbf{0.314} & 0.453 & 0.372 & 0.525 \\
\textsc{Gamma} & 359,298 & \textbf{0.360} & 0.519 & \textbf{0.441} & \textbf{0.578} & 0.310 & \textbf{0.460} & \textbf{0.376} & \textbf{0.526} \\
\bottomrule
\end{tabular}
}
\end{table*}

As shown in Table \ref{tab:parameter_comparison}, models obtained by merely increasing the MLP width and hidden dimensionality do not exhibit the level of overall improvement achieved by \textsc{Gamma}. In contrast, introducing multi-head attention yields a more significant  comprehensive performance boost, and further incorporating the geometric enhancement of the message functions provides an additional layer of gains. These observations indicate that \textsc{Gamma}’s improvements stem from architectural and representational advances rather than from naively scaling parameter count.

\begin{tcolorbox}[takeawaybox]
\textbf{Takeaway 2.} The observed gains cannot be attributed merely to a larger parameter budget. Instead, they highlight the representational advantage of multi-head geometric attention.
\end{tcolorbox}

\subsection{How Do the Fusion Modules Influence Performance?}
Beyond verifying that multi-head geometric attention yields consistent gains, it is essential to understand how the design choices of the fusion module influence performance. In particular, we investigate three key factors: (i) branch fusion mechanism: attention vs. weak or no-attention variants; (ii) attention fusion position: late fusion vs. early fusion; and (iii) branch composition and complementarity: the choice of message functions. By systematically varying these dimensions, we aim to identify the most effective mechanism for integrating heterogeneous relational signals.

\subsubsection{Ablation on Branch Fusion Mechanism}
To examine whether the full attention mechanism is truly necessary, we (i) remove the attention module and directly concatenate the outputs of all branches, (ii) remove the query vector from the attention computation, (iii) exclude the node features from the attention context, and (iv) replace the final feature concatenation with summation. These variants allow us to quantify how each component contributes to the model’s performance.

Table \ref{tab:fusion_mechanism} reports the corresponding results. The largest drops occur for lacking of either attention, query or key (-0.008 or -0.009 total average MRR), indicating that the adaptive weighting driven by query-key interactions is the major contributor to \textsc{Gamma}'s performance gains. By replacing concatenation with summation shows a smaller decrease (-0.006), implying that concatenation mainly improves the representational capacity rather than the adaptive weighting itself. Removing any of these components leads to a noticeable degradation in average MRR, confirming the necessity of the complete attention formulation.

\begin{table*}[htbp]
\caption{Zero-Shot Link Prediction Average Results of \textsc{Gamma}, \textsc{Gamma} without Attention, \textsc{Gamma} without Query in the Attention Computation, \textsc{Gamma} without Key in the Attention Computation and \textsc{Gamma} without Feature Concatenation Grouped by Datasets Generalization Scenarios over 53 KGs.}
\label{tab:fusion_mechanism}
\centering
\resizebox{0.9\textwidth}{!}{
\begin{tabular}{llcccccccc}
\toprule
\multirow{2}{*}{\textbf{Model}} & \multirow{2}{*}{\textbf{Fusion Mechanism}} &
\multicolumn{2}{c}{\makecell{\textbf{Inductive $e,r$} \\ (23 graphs)}} &
\multicolumn{2}{c}{\makecell{\textbf{Inductive $e$} \\ (18 graphs)}} &
\multicolumn{2}{c}{\makecell{\textbf{Transductive} \\ (12 graphs)}} &
\multicolumn{2}{c}{\makecell{\textbf{Total Avg} \\ (53 graphs)}} \\
\cmidrule(lr){3-4} \cmidrule(lr){5-6} \cmidrule(lr){7-8} \cmidrule(lr){9-10}
& & \textbf{MRR} & \textbf{H@10}
  & \textbf{MRR} & \textbf{H@10}
  & \textbf{MRR} & \textbf{H@10}
  & \textbf{MRR} & \textbf{H@10} \\
\midrule
\multirow{4}{*}{\textsc{Gamma}\textsubscript{variants}}
& w/o attention & 0.351 & 0.515 & 0.429 & 0.563 & 0.307 & 0.449 & 0.367 & 0.517 \\
& w/o query & 0.344 & \textbf{0.525} & 0.437 & 0.575 & \textbf{0.313} & 0.457 & 0.368 & \textbf{0.527}\\
& w/o key & 0.355 & 0.519 & 0.431 & 0.572 & 0.298 & 0.441 & 0.368 & 0.520 \\
& w/o concatenation & 0.356 & 0.518 & 0.433 & 0.576 & 0.302 & 0.447 & 0.370 & 0.522 \\
\midrule
\textsc{Gamma} & full attention & \textbf{0.360} & 0.519 & \textbf{0.441} & \textbf{0.578} & 0.310 & \textbf{0.460} & \textbf{0.376} & 0.526 \\
\bottomrule
\end{tabular}
}
\end{table*}

\subsubsection{Ablation on Attention Fusion Position}
We also investigate how the fusion position impacts model performance by training an early fusion version, which merges features after every layer. Table \ref{tab:fusion_position} provides the comparison of the results where early fusion consistently underperforms the late fusion. The pronounced performance drop across almost all scenarios indicates that fusing branch representations too early leads to a loss of predictive performance.

\begin{table*}[htbp]
\caption{Zero-Shot Link Prediction Average Results of \textsc{Gamma} and \textsc{Gamma} with Early Fusion Grouped by Datasets Generalization Scenarios over 53 KGs.}
\label{tab:fusion_position}
\centering
\resizebox{0.9\textwidth}{!}{
\begin{tabular}{llcccccccc}
\toprule
\multirow{2}{*}{\textbf{Model}} & \multirow{2}{*}{\textbf{Fusion Position}} &
\multicolumn{2}{c}{\makecell{\textbf{Inductive $e,r$} \\ (23 graphs)}} &
\multicolumn{2}{c}{\makecell{\textbf{Inductive $e$} \\ (18 graphs)}} &
\multicolumn{2}{c}{\makecell{\textbf{Transductive} \\ (12 graphs)}} &
\multicolumn{2}{c}{\makecell{\textbf{Total Avg} \\ (53 graphs)}} \\
\cmidrule(lr){3-4} \cmidrule(lr){5-6} \cmidrule(lr){7-8} \cmidrule(lr){9-10}
& & \textbf{MRR} & \textbf{H@10}
  & \textbf{MRR} & \textbf{H@10}
  & \textbf{MRR} & \textbf{H@10}
  & \textbf{MRR} & \textbf{H@10} \\
\midrule
\textsc{Gamma}\textsubscript{variant} & early fusion & 0.349 & \textbf{0.522} & 0.432 & \textbf{0.579} & 0.300 & 0.441 & 0.366 & 0.523 \\
\textsc{Gamma} & late fusion & \textbf{0.360} & 0.519 & \textbf{0.441} & 0.578 & \textbf{0.310} & \textbf{0.460} & \textbf{0.376} & \textbf{0.526} \\
\bottomrule
\end{tabular}
}
\end{table*}

\subsubsection{Ablation on Branch Composition and Complementarity}
We further explore the choice of message functions by training multiple \textsc{Gamma} variants with all pairwise combinations. The results are summarized in Table \ref{tab:message_type_full} where the combination of complex and split-complex deliver the largest MRR gains (+0.012) over the best single branch baseline. Interestingly, replacing the split-complex branch with DistMult \cite{DBLP:journals/corr/YangYHGD14a} yields a very similar performance, suggesting strong complementarity between phase sensitive (complex \cite{DBLP:conf/icml/TrouillonWRGB16}) and amplitude or scale oriented multiplicative behaviors (split-complex \cite{DBLP:conf/aaai/PanNLS24} or DistMult). Dual \cite{dong-etal-2024-dual} and DistMult combination also shows a clear gain (+0.006), indicating that first order shear or translation like interactions (dual) complement diagonal multiplicative patterns (DistMult). In contrast, the combination of split-complex and DistMult (+0.005) and the combination of complex and dual (+0.004) exhibit moderate complementarity, while the combination of split-complex and dual (+0.002) is the weakest due to potential higher overlap in the induced feature subspaces. The ablation results reveal that multi-branch message composition consistently improves link prediction performance compared to single branch variants, confirming the effectiveness of our multi-head attention design.

\begin{table*}[htbp]
\caption{Zero-Shot Link Prediction Average Results of \textsc{Ultra} Variants with Different Message Functions, \textsc{Gamma} and \textsc{Gamma} Variants with Different Combinations of Message Functions Grouped by Datasets Generalization Scenarios over 53 KGs.}
\label{tab:message_type_full}
\centering
\resizebox{0.9\textwidth}{!}{
\begin{tabular}{llcccccccc}
\toprule
\multirow{2}{*}{\textbf{Model}} & \multirow{2}{*}{\textbf{Message Type}} &
\multicolumn{2}{c}{\makecell{\textbf{Inductive $e,r$} \\ (23 graphs)}} &
\multicolumn{2}{c}{\makecell{\textbf{Inductive $e$} \\ (18 graphs)}} &
\multicolumn{2}{c}{\makecell{\textbf{Transductive} \\ (12 graphs)}} &
\multicolumn{2}{c}{\makecell{\textbf{Total Avg} \\ (53 graphs)}} \\
\cmidrule(lr){3-4} \cmidrule(lr){5-6} \cmidrule(lr){7-8} \cmidrule(lr){9-10}
& & \textbf{MRR} & \textbf{H@10}
  & \textbf{MRR} & \textbf{H@10}
  & \textbf{MRR} & \textbf{H@10}
  & \textbf{MRR} & \textbf{H@10} \\
\midrule
\multirow{3}{*}{\textsc{Ultra}\textsubscript{variants}}
& complex                         & 0.347 & 0.518 & 0.427 & 0.561 & 0.298 & 0.437 & 0.363 & 0.514 \\
& split-complex                   & 0.343 & 0.518 & 0.435 & 0.572 & 0.299 & 0.440 & 0.364 & 0.519 \\
& dual                            & 0.340 & 0.513 & 0.436 & 0.575 & 0.298 & 0.437 & 0.363 & 0.517 \\
\midrule
\multirow{6}{*}{\textsc{Gamma}\textsubscript{variants}} 
& complex \& dual                 & 0.346 & 0.517 & 0.441 & 0.580 & 0.302 & 0.445 & 0.368 & 0.522 \\
& split-complex \& dual           & 0.347 & 0.517 & 0.433 & 0.568 & 0.303 & 0.436 & 0.366 & 0.516 \\
& complex \& Distmult             & 0.357 & \textbf{0.533} & \textbf{0.442} & \textbf{0.582} & 0.309 & 0.455 & 0.375 & \textbf{0.532} \\
& split-complex \& Distmult       & 0.353 & 0.523 & 0.440 & 0.577 & 0.295 & 0.445 & 0.369 & 0.524 \\
& dual \& Distmult                & 0.348 & 0.513 & 0.439 & 0.580 & \textbf{0.310} & 0.457 & 0.370 & 0.523 \\
\midrule
\textsc{Gamma} & complex \& split-complex       & \textbf{0.360} & 0.519 & 0.441 & 0.578 & \textbf{0.310} & \textbf{0.460} & \textbf{0.376} & 0.526 \\
\bottomrule
\end{tabular}
}
\end{table*}

\begin{tcolorbox}[takeawaybox]
\textbf{Takeaway 3.} The complete attention architecture with late, multi-branch fusion is crucial for achieving the strongest and most expressive performance.
\end{tcolorbox}

\section{Conclusion}
\label{sec:conclusion}
In this work, we introduced \textsc{Gamma}, a structural knowledge graph foundation model that overcomes the limitations of existing approaches by integrating multiple geometric transformations within a unified attention-based framework. Unlike prior models that rely on a single transformation defined in a single geometry and thus introduce structural biases, \textsc{Gamma} exploits the complementarity of geometric message functions to dynamically adapt to the queries with different relational patterns present in the data.

Through extensive evaluation on 53 inductive link prediction benchmarks, \textsc{Gamma} consistently outperforms \textsc{Ultra}, achieving particularly large gains in the fully-inductive setting. These results highlight the importance of multi-geometric reasoning for enabling universal generalization in structural knowledge graph foundation models.

The current regularization settings remain relatively conservative, and the relation model architecture still requires refinement to better accommodate a broader spectrum of relational patterns. We also intend to reduce the model’s parameter size, improve computational efficiency and extend \textsc{Gamma} toward scalable pretraining on large heterogeneous knowledge sources and explore its applicability to downstream tasks beyond link prediction, such as multi-hop reasoning and temporal knowledge graph completion. These directions represent promising avenues for future work.

\section*{Acknowledgments}
The authors gratefully acknowledge the computing time provided on the high-performance computer HoreKa by the National High-Performance Computing Center at KIT (NHR@KIT). This center is jointly supported by the Federal Ministry of Education and Research and the Ministry of Science, Research and the Arts of Baden-Württemberg, as part of the National High-Performance Computing (NHR) joint funding program (https://www.nhr-verein.de/en/our-partners). HoreKa is partly funded by the German Research Foundation (DFG). Mojtaba Nayyeri acknowledges BMBF support through the ATLAS
project (031L0304A).

\bibliographystyle{IEEEtran}
\bibliography{paper}

\section*{Biography}

\begin{IEEEbiographynophoto}{Ling Xin}
is a master's student at the University of Stuttgart, Stuttgart, Baden-Württemberg, Germany. His research interests include machine learning, graph neural networks, and foundation models. Xin received his bachelor's degree in Intelligent Control and Automation from Wuhan Institute of Technology. He is currently pursuing his master’s degree at the University of Stuttgart. Contact him at st176945@stud.uni-stuttgart.de.
\end{IEEEbiographynophoto}

\vspace{11pt}

\begin{IEEEbiographynophoto}{Mojtaba Nayyeri}
received the B.Sc. and M.Sc. degrees in computer engineering from the Ferdowsi University of Mashhad, Mashhad, Iran, in 2011 and 2014, respectively. He received his Ph.D. degree from the Analytics Computing Group, University of Stuttgart, Germany, in 2025, and he is currently a PostDoc researcher at the University of Stuttgart. His current research interests include foundation models, machine learning, knowledge graphs, pattern recognition and the semantic web.
\end{IEEEbiographynophoto}

\vspace{11pt}

\begin{IEEEbiographynophoto}{Zahra Makki Nayeri}
is a Ph.D. candidate at Shahrood University of Technology and currently a visiting researcher at the University of Stuttgart, Germany, where she conducts research on knowledge graph foundation models. Her research focuses on graph representation learning, temporal and dynamic graph neural networks, and machine learning–based modeling of computer networks, with an emphasis on data-driven analysis of large-scale, evolving interaction graphs. During her Master’s studies, she investigated machine learning techniques for fog, edge, and cloud computing environments, with a particular focus on distributed data processing, resource-aware learning, and system-level optimization.
\end{IEEEbiographynophoto}

\vspace{11pt}

\begin{IEEEbiographynophoto}{Professor Steffen Staab}
has studied computer science and computational linguistics at the Universität Erlangen-Nürnberg and at the University of Pennsylvania. He worked in the previous computational linguistics research group at the Universität Freiburg and did his Ph.D. in computer science in the faculty for technology in 1998. He then joined Universität Stuttgart, Institute IAT \& Fraunhofer IAO, before he moved on to the Universität Karlsruhe (now: KIT), where he progressed from project lead, over lecturer and senior lecturer, and did his habilitation in 2002. In 2004, he became a professor for databases and information systems at Universität Koblenz-Landau, where he founded the Institute for Web Science and Technologies (WeST) in 2009 and was head of it until 2020. Since February 2020, he has had a chair for Analytic Computing at the Institute for Parallel and Distributed Systems (and Institute for AI) of Universität Stuttgart. In parallel, he has held a Chair for Web and Computer Science at the University of Southampton since March 2015.
\end{IEEEbiographynophoto}

\vfill

\newpage

{\appendices
\section{Complete Statistics of Datasets}
\label{app:datsets_statistic}
This section provides detailed information about all datasets involved in training and evaluation. We summarize the dataset statistics in four tables. Table \ref{tab:inductiveer_datasets} lists the 23 inductive $e,r$ datasets, where both new entities and relations appear at inference time. Table \ref{tab:inductivee_datasets} presents the 18 inductive $e$ datasets, in which only new entities are introduced during inference while the set of relations remains fixed. Table \ref{tab:transductive_datasets} reports the 15 transductive datasets, which maintain fixed entities and relations across both training and inference. Finally, Table \ref{tab:pattern-stats} reports the frequency of the detected relational pattern types and the number of corresponding triples in test subsets across 56 datasets.

\begin{table*}[htbp]
\caption{23 Inductive $e,r$ Datasets Used in the Experiments. Triples Denotes the Number of Edges in the Corresponding Graph, $Q_{\text{valid}}$ Denotes the Number of Queries in the Validation Set and $Q_{\text{test}}$ Denotes the Number of Queries in the Test Set.}
\label{tab:inductivee_datasets}
\centering
\resizebox{0.8\textwidth}{!}{
\begin{tabular}{llrrrrrrrrrrr}
\toprule
\multirow{2}{*}{\textbf{Dataset}} & \multirow{2}{*}{\textbf{Reference}} &
\multicolumn{3}{c}{\textbf{Training Graph}} &
\multicolumn{4}{c}{\textbf{Validation Graph}} &
\multicolumn{4}{c}{\textbf{Test Graph}} \\
\cmidrule(lr){3-5} \cmidrule(lr){6-9} \cmidrule(lr){10-13}
& & \textbf{Entities} & \textbf{Relations} & \textbf{Triples}
& \textbf{Entities} & \textbf{Relations} & \textbf{Triples} & $\boldsymbol{Q_{\text{valid}}}$
& \textbf{Entities} & \textbf{Relations} & \textbf{Triples} & $\boldsymbol{Q_{\text{test}}}$ \\
\midrule
FB-100 & \cite{pmlr-v202-lee23c}  & 4659  & 134 & 62809  & 2624  & 77  & 6987   & 2329 & 2624  & 77  & 6987   & 2329 \\
FB-50 & \cite{pmlr-v202-lee23c}   & 5190  & 153 & 85375  & 4445  & 205 & 11636  & 3879 & 4445  & 205 & 11636  & 3879 \\
FB-75 & \cite{pmlr-v202-lee23c}   & 4659  & 134 & 62809  & 2792  & 186 & 9316   & 3106 & 2792  & 186 & 9316   & 3106 \\
FB-25 & \cite{pmlr-v202-lee23c}   & 5190  & 163 & 91571  & 4097  & 216 & 17147  & 5716 & 4097  & 216 & 17147  & 5716 \\
WK-100 & \cite{pmlr-v202-lee23c}  & 9784  & 67  & 49875  & 12136 & 37  & 13487  & 4496 & 12136 & 37  & 13487  & 4496 \\
WK-50 & \cite{pmlr-v202-lee23c}   & 12022 & 72  & 82481  & 9328  & 93  & 9672   & 3224 & 9328  & 93  & 9672   & 3225 \\
WK-75 & \cite{pmlr-v202-lee23c}   & 6853  & 52  & 28741  & 2722  & 65  & 3430   & 1143 & 2722  & 65  & 3430   & 1144 \\
WK-25 & \cite{pmlr-v202-lee23c}   & 12659 & 47  & 41873  & 3228  & 74  & 3391   & 1130 & 3228  & 74  & 3391   & 1131 \\
NL-100 & \cite{pmlr-v202-lee23c}  & 1258  & 55  & 7832   & 1709  & 53  & 2378   & 793  & 1709  & 53  & 2378   & 793  \\
NL-75 & \cite{pmlr-v202-lee23c}   & 2607  & 96  & 11058  & 1578  & 116 & 1818   & 606  & 1578  & 116 & 1818   & 607  \\
NL-50 & \cite{pmlr-v202-lee23c}   & 4396  & 106 & 17578  & 2335  & 119 & 2576   & 859  & 2335  & 119 & 2576   & 859  \\
NL-25 & \cite{pmlr-v202-lee23c}   & 4396  & 106 & 17578  & 2146  & 120 & 2230   & 744  & 2146  & 120 & 2230   & 744 \\
NL-0 & \cite{pmlr-v202-lee23c}    & 1814  & 134 & 7796   & 2026  & 112 & 2287   & 763  & 2026  & 112 & 2287   & 763  \\
\textsc{WikiTopics-MT1:Tax} & \cite{DBLP:journals/corr/abs-2307-06046}    & 10000 & 10  & 17178  & 10000 & 10  & 17178  & 1908 & 10000 & 9   & 16526  & 1834 \\
\textsc{WikiTopics-MT1:Health} & \cite{DBLP:journals/corr/abs-2307-06046} & 10000 & 7   & 14371  & 10000 & 7   & 14371  & 1596 & 10000 & 7   & 14110  & 1566 \\
\textsc{WikiTopics-MT2:Org} & \cite{DBLP:journals/corr/abs-2307-06046}    & 10000 & 10  & 23233  & 10000 & 10  & 23233  & 2581 & 10000 & 11  & 21976  & 2441 \\
\textsc{WikiTopics-MT2:Sci} & \cite{DBLP:journals/corr/abs-2307-06046}    & 10000 & 16  & 16471  & 10000 & 16  & 16471  & 1830 & 10000 & 16  & 14852  & 1650 \\
\textsc{WikiTopics-MT3:Art} & \cite{DBLP:journals/corr/abs-2307-06046}    & 10000 & 45  & 27262  & 10000 & 45  & 27262  & 3026 & 10000 & 45  & 28023  & 3113 \\
\textsc{WikiTopics-MT3:Infra} & \cite{DBLP:journals/corr/abs-2307-06046}  & 10000 & 24  & 21990  & 10000 & 24  & 21990  & 2443 & 10000 & 27  & 21646  & 2405 \\
\textsc{WikiTopics-MT4:Sci} & \cite{DBLP:journals/corr/abs-2307-06046}    & 10000 & 42  & 12576  & 10000 & 42  & 12576  & 1397 & 10000 & 42  & 12516  & 1388 \\
\textsc{WikiTopics-MT4:Health} & \cite{DBLP:journals/corr/abs-2307-06046} & 10000 & 21  & 15539  & 10000 & 21  & 15539  & 1725 & 10000 & 20  & 15337  & 1703 \\
\textsc{MetaFam}   & \cite{DBLP:journals/corr/abs-2307-06046} & 1316  & 28  & 13821  & 1316  & 28  & 13821  & 590  & 656   & 28  & 7257   & 184  \\
FBNELL & \cite{DBLP:journals/corr/abs-2307-06046}    & 4636  & 100 & 10275  & 4636  & 100 & 10275  & 1055 & 4752  & 183 & 10685  & 597  \\
\bottomrule
\end{tabular}
}
\end{table*}

\begin{table*}[htbp]
\caption{18 Inductive $e$ Datasets Used in the Experiments. Triples Denotes the Number of Edges in the Corresponding Graph, $Q_{\text{valid}}$ Denotes the Number of Queries in the Validation Set and $Q_{\text{test}}$ Denotes the Number of Queries in the Test Set.}
\label{tab:inductiveer_datasets}
\centering
\resizebox{0.8\textwidth}{!}{
\begin{tabular}{l l r r r r r r r r r r}
\toprule
\multirow{2}{*}{\textbf{Dataset}} & \multirow{2}{*}{\textbf{Reference}} & \multirow{2}{*}{\textbf{Relations}} &
\multicolumn{2}{c}{\textbf{Training Graph}} &
\multicolumn{3}{c}{\textbf{Validation Graph}} &
\multicolumn{3}{c}{\textbf{Test Graph}} \\
\cmidrule(lr){4-5} \cmidrule(lr){6-8} \cmidrule(lr){9-11}
& & & \textbf{Entities} & \textbf{Triples} 
  & \textbf{Entities} & \textbf{Triples} & $\boldsymbol{Q_{\text{valid}}}$
  & \textbf{Entities} & \textbf{Triples} & $\boldsymbol{Q_{\text{test}}}$ \\
\midrule
WN18RR:v1 & \cite{DBLP:conf/icml/TeruDH20}   & 9   & 2746  & 5410   & 2746  & 5410   & 630   & 922   & 1618   & 373  \\
WN18RR:v2 & \cite{DBLP:conf/icml/TeruDH20}   & 10  & 6954  & 15262  & 6954  & 15262  & 1838  & 2757  & 4011   & 852  \\
WN18RR:v3 & \cite{DBLP:conf/icml/TeruDH20}   & 11  & 12078 & 25901  & 12078 & 25901  & 3097  & 5084  & 6327   & 1143 \\
WN18RR:v4 & \cite{DBLP:conf/icml/TeruDH20}   & 9   & 3861  & 7940   & 3861  & 7940   & 934   & 7084  & 12334  & 2823 \\
FB15k-237:v1 & \cite{DBLP:conf/icml/TeruDH20}   & 180 & 1594  & 4245   & 1594  & 4245   & 489   & 1093  & 1993   & 411  \\
FB15k-237:v2 & \cite{DBLP:conf/icml/TeruDH20}   & 200 & 2608  & 9739   & 2608  & 9739   & 1166  & 1660  & 4145   & 947  \\
FB15k-237:v3 & \cite{DBLP:conf/icml/TeruDH20}   & 215 & 3668  & 17986  & 3668  & 17986  & 2194  & 2501  & 7406   & 1731 \\
FB15k-237:v4 & \cite{DBLP:conf/icml/TeruDH20}   & 219 & 4707  & 27203  & 4707  & 27203  & 3352  & 3051  & 11714  & 2840 \\
NELL-995:v1 & \cite{DBLP:conf/icml/TeruDH20} & 14  & 3103  & 4687   & 3103  & 4687   & 414   & 225   & 833    & 201  \\
NELL-995:v2 & \cite{DBLP:conf/icml/TeruDH20} & 88  & 2564  & 8219   & 2564  & 8219   & 922   & 2086  & 4586   & 935  \\
NELL-995:v3 & \cite{DBLP:conf/icml/TeruDH20} & 142 & 4647  & 16393  & 4647  & 16393  & 1851  & 3566  & 8048   & 1620 \\
NELL-995:v4 & \cite{DBLP:conf/icml/TeruDH20} & 76  & 2092  & 7546   & 2092  & 7546   & 876   & 2795  & 7073   & 1447 \\
ILPC22-S & \cite{DBLP:journals/corr/abs-2203-01520} & 48 & 10230 & 78616  & 6653  & 20960  & 2906  & 6653  & 20960  & 2902 \\
ILPC22-L & \cite{DBLP:journals/corr/abs-2203-01520} & 65 & 46626 & 202446 & 29246 & 77044  & 10179 & 29246 & 77044  & 10184 \\
Hamaguchi-BM:1k & \cite{DBLP:conf/ijcai/HamaguchiOSM17}   & 11 & 36237 & 93364  & 36311 & 93364  & 1771  & 9899  & 18638  & 476  \\
Hamaguchi-BM:3k & \cite{DBLP:conf/ijcai/HamaguchiOSM17}   & 11 & 32118 & 71097  & 32250 & 71097  & 1201  & 19218 & 38285  & 1349 \\
Hamaguchi-BM:5k & \cite{DBLP:conf/ijcai/HamaguchiOSM17}  & 11 & 28601 & 57601  & 28744 & 57601  & 900   & 23792 & 48425  & 2124 \\
INDIGO-BM & \cite{NEURIPS2021_0fd600c9}      & 229 & 12721 & 121601 & 12797 & 121601 & 14121 & 14775 & 250195 & 14904 \\
\bottomrule
\end{tabular}
}
\end{table*}

\begin{table*}[htbp]
\caption{15 Transductive Datasets Used in the Experiments. Train, Valid, Test Denote the Number of Triples in the Respective Set. Task Denotes the Prediction Task at Inference: h/t is Predicting both Heads and Tails, t is only
Predicting Tails. }
\label{tab:transductive_datasets}
\centering
\tiny
\resizebox{0.8\textwidth}{!}{
\begin{tabular}{l l r r r r r l}
\hline
\textbf{Dataset} & \textbf{Reference} & \textbf{Entities} & \textbf{Relations} & \textbf{Train} & \textbf{Valid} & \textbf{Test} & \textbf{Task}  \\
\hline
WN18RR           & \cite{DBLP:conf/aaai/DettmersMS018}       & 40943  & 11   & 86835   & 3034   & 3134   & h/t \\
FB15k-237         & \cite{toutanova2015observed}     & 14541  & 237  & 272115  & 17535  & 20466  & h/t \\
CoDEx-M     & \cite{safavi-koutra-2020-codex}      & 17050  & 51   & 185584  & 10310  & 10311  & h/t \\
CoDEx-S      & \cite{safavi-koutra-2020-codex}      & 2034   & 42   & 32888   & 1827   & 1828   & h/t \\
CoDEx-L      & \cite{safavi-koutra-2020-codex}      & 77951  & 69   & 551193  & 30622  & 30622  & h/t \\
NELL-995          & \cite{xiong-etal-2017-deeppath}          & 74536  & 200  & 149678  & 543    & 2818   & h/t \\
YAGO310          & \cite{Mahdisoltani2015YAGO3AK}   & 123182 & 37   & 1079040 & 5000   & 5000   & h/t \\
WD-singer         & \cite{lv-etal-2020-dynamic}             & 10282  & 135  & 16142   & 2163   & 2203   & h/t \\
NELL23K          & \cite{lv-etal-2020-dynamic}             & 22925  & 200  & 25445   & 4961   & 4952   & h/t \\
FB15k-237-10\%    & \cite{lv-etal-2020-dynamic}             & 11512  & 237  & 27211   & 15624  & 18150  & t \\
FB15k-237-20\%     & \cite{lv-etal-2020-dynamic}             & 13166  & 237  & 54423   & 16963  & 19776  & t \\
FB15k-237-50\%     & \cite{lv-etal-2020-dynamic}             & 14149  & 237  & 136057  & 17449  & 20324  & t \\
DB100K      & \cite{ding-etal-2018-improving}           & 99604  & 470  & 597572  & 50000  & 50000  & h/t \\
Aristo-V4         & \cite{DBLP:conf/akbc/ChenM0S21}           & 44949  & 1605 & 242567  & 20000  & 20000  & h/t \\
ConceptNet-100K   & \cite{DBLP:conf/aaai/MalaviyaBBC20}       & 78334  & 34   & 100000  & 1200   & 1200   & h/t \\
\hline
\end{tabular}
}
\end{table*}

\begin{table}[htbp]
\caption{Pattern Statistics Across 56 Datasets. Pattern Denotes The Frequency of The Detected Pattern Types and Triples Denotes the Number of Corresponding Triples in Test Subsets}
\label{tab:pattern-stats}
\centering
\resizebox{\linewidth}{!}{
\begin{tabular}{lrrrrrr}
\toprule
\multirow{2}{*}{\textbf{Pattern Datasets}} &
\multicolumn{2}{c}{\textbf{Symmetry}} &
\multicolumn{2}{c}{\textbf{Anti-symmetry}} &
\multicolumn{2}{c}{\textbf{Composition}} \\
\cmidrule(lr){2-3} \cmidrule(lr){4-5} \cmidrule(lr){6-7}
& \textbf{Patterns} & \textbf{Triple} 
& \textbf{Patterns} & \textbf{Triple}
& \textbf{Patterns} & \textbf{Triple} \\
\midrule
FB-100 & 3 & 0 & 180 & 2258 & 92 & 557 \\
FB-75 & 3 & 43 & 186 & 2874 & 122 & 1969 \\
FB-50 & 3 & 150 & 193 & 3150 & 118 & 2433 \\
FB-25 & 4 & 173 & 199 & 4786 & 131 & 3693 \\
WK-100 & 3 & 0 & 96 & 4138 & 14 & 443 \\
WK-75 & 4 & 121 & 77 & 961 & 21 & 471 \\
WK-50 & 3 & 61 & 105 & 2608 & 20 & 666 \\
WK-25 & 2 & 33 & 81 & 947 & 10 & 477 \\
NL-100 & 0 & 0 & 91 & 460 & 14 & 197 \\
NL-75 & 1 & 10 & 137 & 185 & 24 & 197 \\
NL-50 & 0 & 0 & 132 & 586 & 28 & 273 \\
NL-25 & 2 & 1 & 130 & 549 & 22 & 162 \\
NL-0 & 0 & 0 & 112 & 469 & 34 & 283 \\
\textsc{WikiTopics-MT1:Tax} & 1 & 19 & 9 & 1815 & 0 & 0 \\
\textsc{WikiTopics-MT1:Health} & 1 & 206 & 6 & 1360 & 0 & 0 \\
\textsc{WikiTopics-MT2:Org} & 0 & 0 & 10 & 2422 & 2 & 32 \\
\textsc{WikiTopics-MT2:Sci} & 1 & 54 & 15 & 1596 & 1 & 0 \\
\textsc{WikiTopics-MT3:Art} & 0 & 0 & 52 & 3113 & 5 & 1422 \\
\textsc{WikiTopics-MT3:Infra} & 2 & 759 & 23 & 1559 & 1 & 15 \\
\textsc{WikiTopics-MT4:Sci} & 0 & 0 & 42 & 1383 & 0 & 0 \\
\textsc{WikiTopics-MT4:Health} & 2 & 234 & 18 & 1196 & 1 & 103 \\
\textsc{MetaFam} & 0 & 0 & 22 & 184 & 26 & 184 \\
FBNELL & 0 & 0 & 161 & 267 & 19 & 100 \\
\midrule
WN18RR:v1 & 0 & 0 & 5 & 55 & 0 & 0 \\
WN18RR:v2 & 0 & 0 & 6 & 207 & 0 & 0 \\
WN18RR:v3 & 0 & 0 & 7 & 792 & 0 & 0 \\
WN18RR:v4 & 1 & 3 & 5 & 750 & 0 & 0 \\
FB15k-237:v1 & 6 & 0 & 157 & 339 & 29 & 115 \\
FB15k-237:v2 & 1 & 1 & 176 & 751 & 32 & 260 \\
FB15k-237:v3 & 0 & 0 & 191 & 1408 & 40 & 529 \\
FB15k-237:v4 & 1 & 0 & 194 & 2294 & 37 & 775 \\
NELL-995:v1 & 0 & 0 & 13 & 151 & 0 & 0 \\
NELL-995:v2 & 0 & 0 & 70 & 486 & 15 & 145 \\
NELL-995:v3 & 1 & 1 & 113 & 556 & 26 & 490 \\
NELL-995:v4 & 0 & 0 & 64 & 610 & 21 & 368 \\
ILPC22-S & 0 & 0 & 44 & 2765 & 8 & 431 \\
ILPC22-L & 0 & 0 & 60 & 9824 & 17 & 6317 \\
Hamaguchi-BM:1k & 0 & 0 & 10 & 476 & 0 & 0 \\
Hamaguchi-BM:3k & 0 & 0 & 10 & 1349 & 0 & 0 \\
Hamaguchi-BM:5k & 0 & 0 & 10 & 2123 & 0 & 0 \\
INDIGO-BM & 3 & 5 & 197 & 13143 & 116 & 8914 \\
\midrule
CoDEx-S & 3 & 295 & 38 & 1532 & 10 & 360 \\
CoDEx-L & 0 & 0 & 64 & 29610 & 15 & 9697 \\
NELL-995 & 0 & 0 & 160 & 2818 & 30 & 745 \\
YAGO310 & 2 & 19 & 30 & 4839 & 3 & 323 \\
WD-singer & 1 & 0 & 121 & 1914 & 47 & 1536 \\
NELL23K & 0 & 0 & 168 & 3526 & 22 & 801 \\
FB15k-237-10\% & 3 & 50 & 208 & 17557 & 99 & 10962 \\
FB15k-237-30\% & 3 & 68 & 205 & 19128 & 112 & 12365 \\
FB15k-237-50\% & 3 & 73 & 205 & 19667 & 132 & 12662 \\
DB100K & 2 & 10 & 433 & 40341 & 177 & 38999 \\
Aristo-V4 & 0 & 0 & 1495 & 18963 & 1435 & 19934 \\
ConceptNet-100K & 0 & 0 & 32 & 1199 & 6 & 693 \\
\midrule
WN18RR & 3 & 1116 & 7 & 1962 & 1 & 172 \\
FB15k-237 & 3 & 74 & 205 & 19804 & 144 & 13606 \\
CoDEx-M & 2 & 340 & 47 & 9898 & 7 & 1875 \\
\bottomrule
\end{tabular}
}
\end{table}

\section{Implementation Details}
\label{app:experiment setup details}
\subsection{Model Architecture Modifications}
We build upon the open-source PyG implementation of \textsc{Ultra}\footnotemark[1], extending its original layer design to incorporate additional message functions, including split-complex, dual, mobius, mobius+, splitmobius, and transrotate. The dimensionality of relation embeddings is dynamically adjusted to match the requirements of each message function.
\footnotetext[1]{Open-source code of \textsc{Ultra}: \url{https://github.com/DeepGraphLearning/ULTRA}}
Furthermore, we extend EntityNBFNet into a multi-branch attention fusion architecture, where each message function maintains an independent stack of \textsc{NBFNet} layers and performs forward propagation separately. All branches execute Bellman-Ford iterations in parallel to produce multi-channel representations. The final feature is obtained via an attention fusion mechanism: The query is projected to form a context vector, and each branch feature is projected to form keys in the same space. Relevance is computed via cosine similarity with a temperature scaling, followed by a softmax to obtain attention weights. The attention can be mixed with a uniform distribution to avoid collapse, an entropy term is computed for regularization, and attention dropout is applied. The weighted branch features are then concatenated and passed through an MLP for final scoring.

\subsection{Hyperparameters and Attention Projection}
For the newly added \textsc{NBFNet} branches, we adopt the same hyperparameter configuration as in \textsc{Ultra} \cite{DBLP:conf/iclr/0001YM0Z24} to ensure comparability. Both the query and the branch specific keys are projected into the attention space using a single linear transformation layer. To accommodate the concatenated feature, we increase the input and hidden dimensions of the MLP used for final scoring. Details are presented in Table \ref{tab:gamma_hparams}.

\subsection{Pretraining Procedure}
Following \textsc{Ultra} \cite{DBLP:conf/iclr/0001YM0Z24}, we perform multi-graph pretraining on a collection of heterogeneous knowledge graphs. At each step, a graph is sampled in proportion to its number of edges, and a mini-batch of target edges is used to construct positive and negative triplets. We employ binary cross-entropy loss for training the model to correctly predict the tail entity, attention entropy regularization loss to prevent the attention mechanism from collapsing onto a single branch, adversarial negative sampling to improve the model’s discriminative ability, and adopt gradient accumulation for large-batch optimization. The pretraining process is distributed across 4 GPUs using \texttt{DistributedDataParallel} and presented in Algorithm \ref{alg:pretraining}.

\begin{algorithm}[htbp]
\caption{Multi-Graph Pretraining with \textsc{\textsc{Gamma}}}
\label{alg:pretraining}
\begin{algorithmic}[0]
\Require Training graphs $\{\mathcal G_1,\dots,\mathcal G_N\}$, validation graphs $\{\mathcal G^{\mathrm{val}}\}$, encoder $f_\theta$ (\textsc{Gamma}), epochs $T$, batch size $B$, negatives $n_{\mathrm{neg}}$, accumulation $n_{\mathrm{acc}}$, adversarial temperature $\tau$, aux weight $\lambda_{\mathrm{aux}}$ (default $=1$)
\Ensure Pretrained parameters $\theta^\star$
\State Initialize $\theta$, optimizer $\mathcal O$
\For{epoch $=1$ to $T$}
  \For{each mini-batch}
    \State Sample graph $\mathcal G \sim p(\mathcal G)\propto|\mathcal E_{\mathcal G}|$
    \State Sample $B$ positive triples $\{(h_i,r_i,t_i)\}_{i=1}^B$ from $\mathcal G$
    \State Build negatives by corrupting head/tail:\\
      \quad $\mathcal N_i=\{(h'_i,r_i,t_i)\}\cup\{(h_i,r_i,t'_i)\}$ with $|\mathcal N_i|=n_{\mathrm{neg}}$
    \State Compute logits \emph{and} auxiliary signals via\\
      \quad $(S,\mathcal A)\leftarrow f_\theta(\mathcal G; \{(h_i,r_i,[t_i;\mathcal N_i])\}_{i=1}^B)$,\\
      \quad $S\in\mathbb R^{B\times(1+n_{\mathrm{neg}})}$
    \State Targets $Y_{i,1}=1$, $Y_{i,2:}=0$
    \State Negative weights: $W_{i,1}=1$\\
      \quad $W_{i,2:}=\begin{cases}
      \mathrm{softmax}(S_{i,2:}/\tau), & \tau>0\\
      \mathbf 1/n_{\mathrm{neg}}, & \text{otherwise}
      \end{cases}$
    \State Main loss:\\
      \quad $\displaystyle
      \mathcal L_{\mathrm{main}}=\frac{1}{B}\sum_{i=1}^B
      \frac{\sum_{j} W_{i,j}\cdot \mathrm{BCEWithLogits}(S_{i,j},Y_{i,j})}
           {\sum_{j} W_{i,j}}$
    \State Auxiliary loss:\\
    \quad $\;\mathcal L_{\mathrm{aux}} \leftarrow \mathcal A.\mathrm{aux\_loss}$ if available, else $0$
    \State Total loss:\\
      \quad $\tilde{\mathcal L}=\mathcal L_{\mathrm{main}}+\lambda_{\mathrm{aux}}\cdot \mathcal L_{\mathrm{aux}}$
    \State Accumulate gradients on $\tilde{\mathcal L}/n_{\mathrm{acc}}$ and update $\theta$ every $n_{\mathrm{acc}}$ steps
  \EndFor
  \If{epoch is a checkpoint epoch}
    \State Save checkpoint; evaluate on $\{\mathcal G^{\mathrm{val}}\}$ with filtered ranking
  \EndIf
\EndFor
\State Load best checkpoint by average validation MRR and \Return $\theta^\star$
\end{algorithmic}
\end{algorithm}

\begin{table}[htbp]
\caption{Hyperparameter Settings for \textsc{\textsc{Gamma}} Pre-Training. 
GNN$_r$ is a GNN over the Relation Graph $\mathcal{G}_r$, 
and GNN$_e$ is a GNN over the Entity Graph $\mathcal{G}$.}
\label{tab:gamma_hparams}
\centering
\renewcommand{\arraystretch}{1.05}
\setlength{\tabcolsep}{4pt}
\begin{tabularx}{\columnwidth}{@{}lXX@{}}
\toprule
 & \textbf{Hyperparameter} & \textbf{\textsc{Gamma} pre-training} \\
\midrule
\multirow{4}{*}{\textbf{GNN$_r$}} 
& \# layers & 6 \\
& hidden dim & 64 \\
& message & DistMult \\
& aggregation & sum \\
\midrule
\multirow{7}{*}{\textbf{GNN$_e$}} 
& \# layers per branch & 6 \\
& hidden dim per branch & 64 \\
& message & Complex \& Split-complex \\
& aggregation & sum \\
& $g(\cdot)$ & 2-layer MLP \\
& attention fusion & temperature-scaled cosine attention \\
& \# branches & 2 \\
\midrule
\multirow{6}{*}{\textbf{Learning}} 
& optimizer & AdamW \\
& learning rate & 5e$^{-4}$ \\
& training steps & 200k \\
& adv temperature & 1 \\
& \# negatives & 128 \\
& batch size & 64 \\
& Training graph mixture & FB15k-237,WN18RR,CoDEx-M \\
\bottomrule
\end{tabularx}
\end{table}

\section{Complexity Analysis}
\label{app:complexity}

As discussed in \textsc{Ultra} \cite{DBLP:conf/iclr/0001YM0Z24}, the overall time complexity is mainly dominated by the entity-level graph encoder $\mathrm{GNN}_e$. 
Our model follows the same design and adopts \textsc{NBFNet} \cite{DBLP:conf/nips/ZhuZXT21} as $\mathrm{GNN}_e$. 
In this case, for a single message-passing branch, the time complexity per layer is generally linear in the number of edges, i.e., 
$\mathcal{O}(|\mathcal{E}|d + |\mathcal{V}|d^{2})$. 
With $T$ layers and $K$ parallel message branches, the overall time complexity of the message passing stage becomes 
$\mathcal{O}\!\left(KT(|\mathcal{E}|d + |\mathcal{V}|d^{2})\right)$. 
After message propagation, the multi-branch attention fusion introduces an additional 
$\mathcal{O}(K|\mathcal{V}|F^{2})$ cost, where $F$ denotes the feature dimension of each branch output. 
The final two-layer MLP further contributes $\mathcal{O}(M(KF)^{2})$ complexity for scoring, 
where $M$ is the number of candidate tail entities per query (including negatives). 
In practice, the first term dominates when $|\mathcal{E}| \gg |\mathcal{V}|d$, 
and the overall complexity remains approximately linear in the number of edges.

The memory complexity of a single-branch \textsc{NBFNet} \cite{DBLP:conf/nips/ZhuZXT21} is $\mathcal{O}(T|\mathcal{V}|d)$ due to its kernelized implementation of relational message passing. 
Our two-branch variant requires storing intermediate states and attention projections for each branch, 
leading to $\mathcal{O}(KT|\mathcal{V}|d + K|\mathcal{V}|F)$ memory usage, which is still linear in the number of nodes. 

\section{Detailed Results on each Dataset and Parameter Count Analysis}
\label{app:results}
In this section, we present a comprehensive comparison between \textsc{Ultra} \cite{DBLP:conf/iclr/0001YM0Z24} and \textsc{Gamma} in Table \ref{tab:full_results}. The results include these two models’ performances on the three pre-training datasets and their zero-shot inference outcomes across all test datasets. 

\begin{table}[htbp]
\caption{Zero-shot Link Prediction Performance of \textsc{Ultra} and \textsc{Gamma} on 23 Inductive $e,r$ Datasets, 18 Inductive $e$ Datasets and 12 Transductive Datasets along with Test Results on 3 Pretrained Datasets.}
\label{tab:full_results}
\centering
\resizebox{\linewidth}{!}{
\begin{tabular}{lcccccc}
\toprule
\multirow{2}{*}{\textbf{Dataset}} &
\multicolumn{2}{c}{\textbf{\textsc{Ultra}}} &
\multicolumn{2}{c}{\textbf{\textsc{Gamma}}} \\
\cmidrule(lr){2-3} \cmidrule(lr){4-5}
& \textbf{MRR} & \textbf{Hits@10} & \textbf{MRR} & \textbf{Hits@10} \\
\midrule
\multicolumn{5}{c}{\textit{Inductive $e,r$}} \\
\midrule
FB-100 & 0.435 & 0.627 & \textbf{0.443} & \textbf{0.634} \\
FB-75  & \textbf{0.397} & 0.592 & 0.395 & \textbf{0.599} \\
FB-50  & \textbf{0.335} & 0.536 & 0.332 & \textbf{0.538} \\
FB-25  & 0.389 & 0.636 & \textbf{0.391} & \textbf{0.640} \\
WK-100 & 0.176 & 0.286 & \textbf{0.182} & \textbf{0.299} \\
WK-75  & 0.362 & 0.499 & \textbf{0.392} & \textbf{0.538} \\
WK-50  & 0.148 & 0.291 & \textbf{0.160} & \textbf{0.317} \\
WK-25  & 0.290 & 0.491 & \textbf{0.307} & \textbf{0.498} \\
NL-100 & 0.441 & 0.636 & \textbf{0.469} & \textbf{0.666} \\
NL-75  & 0.308 & 0.484 & \textbf{0.357} & \textbf{0.530} \\
NL-50  & 0.367 & 0.539 & \textbf{0.401} & \textbf{0.570} \\
NL-25  & 0.372 & 0.527 & \textbf{0.388} & \textbf{0.540} \\
NL-0   & 0.353 & 0.528 & \textbf{0.368} & \textbf{0.564} \\
\textsc{WikiTopics-MT1:Tax}    & 0.233 & 0.311 & \textbf{0.260} & \textbf{0.339} \\
\textsc{WikiTopics-MT1:Health} & 0.311 & \textbf{0.405} & \textbf{0.348} & 0.401 \\
\textsc{WikiTopics-MT2:Org}    & 0.087 & 0.144 & \textbf{0.096} & \textbf{0.159} \\
\textsc{WikiTopics-MT3:Sci}    & 0.245 & 0.349 & \textbf{0.280} & \textbf{0.368} \\
\textsc{WikiTopics-MT3:Art}    & 0.261 & 0.419 & \textbf{0.282} & \textbf{0.443} \\
\textsc{WikiTopics-MT3:Infra}  & 0.636 & 0.783 & \textbf{0.650} & \textbf{0.785} \\
\textsc{WikiTopics-MT4:Sci}    & 0.286 & 0.448 & \textbf{0.298} & \textbf{0.463} \\
\textsc{WikiTopics-MT4:Health} & 0.611 & \textbf{0.750} & \textbf{0.650} & \textbf{0.750} \\
\textsc{MetaFam}               & \textbf{0.442} & \textbf{0.842} & 0.342 & 0.660 \\
FBNELL                         & 0.477 & 0.625 & \textbf{0.480} & \textbf{0.645} \\
\midrule
\multicolumn{5}{c}{\textit{Inductive $e$}} \\
\midrule
WN18RR:v1       & 0.421 & 0.599 & \textbf{0.682} & \textbf{0.783} \\
WN18RR:v2       & 0.632 & \textbf{0.765} & \textbf{0.672} & 0.764 \\
WN18RR:v3       & 0.388 & 0.511 & \textbf{0.397} & \textbf{0.522} \\
WN18RR:v4       & 0.592 & 0.712 & \textbf{0.610} & \textbf{0.713} \\\textbf{}
FB15k-237:v1    & 0.497 & 0.652 & \textbf{0.500} & \textbf{0.661} \\
FB15k-237:v2    & 0.507 & \textbf{0.696} & \textbf{0.511} & 0.695 \\
FB15k-237:v3    & \textbf{0.493} & \textbf{0.657} & 0.491 & 0.646 \\
FB15k-237:v4    & 0.480 & \textbf{0.674} & \textbf{0.486} & \textbf{0.674} \\
NELL-995:v1     & 0.683 & 0.866 & \textbf{0.791} & \textbf{0.915} \\
NELL-995:v2     & 0.509 & 0.702 & \textbf{0.527} & \textbf{0.724} \\
NELL-995:v3     & 0.512 & 0.693 & \textbf{0.524} & \textbf{0.700} \\
NELL-995:v4     & 0.497 & 0.716 & \textbf{0.506} & \textbf{0.725 }\\
ILPC22-S        & 0.297 & 0.450 & \textbf{0.305} & \textbf{0.452} \\
ILPC22-L        & 0.303 & 0.425 & \textbf{0.308} & \textbf{0.428} \\
Hamaguchi-BM:1k & 0.070 & 0.130 & \textbf{0.074} & \textbf{0.146} \\
Hamaguchi-BM:3k & 0.049 & 0.093 & \textbf{0.058} & \textbf{0.108} \\
Hamaguchi-BM:5k & 0.044 & 0.084 & \textbf{0.053} & \textbf{0.102} \\
INDIGO-BM       & 0.442 & 0.647 & \textbf{0.448} & \textbf{0.653} \\
\midrule
\multicolumn{5}{c}{\textit{Transductive}} \\
\midrule
CoDEx-S         & 0.477 & 0.659  & \textbf{0.480} & \textbf{0.668} \\
CoDEx-L         & 0.338 & 0.468  & \textbf{0.341} & \textbf{0.475} \\
NELL-995        & \textbf{0.498} & \textbf{0.612} & 0.475 & 0.602 \\
YAGO310         & \textbf{0.509} & \textbf{0.668} & 0.412 & 0.586 \\
WD-singer       & 0.362 & 0.473  & \textbf{0.383} & \textbf{0.504} \\
NELL23K         & 0.234 & 0.392  & \textbf{0.241} & \textbf{0.418} \\
FB15k-237-10\%  & 0.156 & 0.276  & \textbf{0.161} & \textbf{0.280} \\
FB15k-237-20\%  & 0.183 & 0.314  & \textbf{0.184} & \textbf{0.315} \\
FB15k-237-50\%  & 0.231 & 0.401  & \textbf{0.235} & \textbf{0.406} \\
DB100K          & 0.412 & 0.579  & \textbf{0.420} & \textbf{0.587} \\
Aristo-V4       & 0.203 & 0.299  & \textbf{0.206} & \textbf{0.315} \\
ConceptNet-100K & 0.112 & 0.206  & \textbf{0.186} & \textbf{0.361} \\
\midrule
\multicolumn{5}{c}{\textit{Pretrained}} \\
\midrule
WN18RR    & 0.503 & 0.623 & \textbf{0.505} & \textbf{0.632} \\
FB15k-237 & 0.371 & 0.568 & \textbf{0.377} & \textbf{0.571} \\
CoDEx-M   & 0.375 & 0.531 & \textbf{0.376} & \textbf{0.532} \\
\bottomrule
\end{tabular}
}
\end{table}
}

\end{document}